\documentclass[10pt,journal,compsoc]{IEEEtran}
\ifCLASSOPTIONcompsoc
  \usepackage[nocompress]{cite}
\else
  \usepackage{cite}
\fi
\ifCLASSINFOpdf
   \usepackage[pdftex]{graphicx}
\else
   \usepackage[dvips]{graphicx}
\fi
\usepackage{amsmath}
\usepackage{amsfonts,amssymb}
\usepackage{mathrsfs}
\usepackage{booktabs}
\usepackage{multirow}
\usepackage{makecell}
\newtheorem{myDef}{Definition}
\newtheorem{myProp}{Property}
\usepackage{pgfplots}

\usepackage{algorithm}
\usepackage{algpseudocode}
\usepackage{subcaption}
\usepackage{cleveref}
\usepackage{url}
\begin{document}
\title{Outlining and Filling: Hierarchical Query Graph Generation for Answering Complex Questions over Knowledge Graphs}
% one-size-fits-all method a daunting prospect. For instance, compsoc 

\author{Yongrui~Chen,%~\IEEEmembership{Member,~IEEE,}
        Huiying~Li,%~\IEEEmembership{Fellow,~OSA,}
        Guilin Qi,
        Tianxing Wu,
        and~Tenggou~Wang%~\IEEEmembership{Life~Fellow,~IEEE}% <-this % stops a space
        
\IEEEcompsocitemizethanks{\IEEEcompsocthanksitem Y. Chen, H. Li, G. Qi, T. Wu, and T. Wang are with 
School of Computer Science and Engineering, 
Southeast University, Nanjing 210096, China.\protect\\
E-mail: \{yrchen, huiyingli, gqi, tianxingwu, wangtenggou\}@seu.edu.cn}% <-this % stops an unwanted space
}


\markboth{Journal of \LaTeX\ Class Files,~Vol.~14, No.~8, August~2015}%
{Chen \MakeLowercase{\textit{et al.}}: Outlining and Filling: Hierarchical Query Graph Generation for Answering Complex Questions over Knowledge Graph}



\IEEEtitleabstractindextext{%
\begin{abstract}
% query graph construction aims to build correct executable SPARQL over the knowledge graph for answering natural language questions. Although recent methods perform well by NN-based query graph ranking, more complex questions bring three new challenges: complicated SPARQL syntax, huge search space for ranking, and noisy query graphs with local ambiguity. This paper handles these challenges. Initially, we regard common complicated SPARQL syntax as the sub-graphs comprising of vertices and edges and propose a new unified query graph grammar to adapt them. Subsequently, we propose a new two-stage method to build query graphs. In the first stage, the top-$k$ related instances (entities, relations, etc.) are collected by simple strategies, as the candidate instances. In the second stage, a graph generation model performs hierarchical generation.  It first outlines a graph structure whose vertices and edges are empty slots, and then fills the appropriate instances into the slots, thereby completing the query graph. Our method decomposes the unbearable search space of entire query graphs into affordable sub-spaces of operations, meanwhile, leverages the global structural information to eliminate local ambiguity. The experimental results demonstrate that our method greatly improves state-of-the-art on the hardest KGQA benchmarks and has an excellent performance on complex questions.

Query graph construction aims to construct the correct executable SPARQL on the KG to answer natural language questions. Although recent methods have achieved good results using neural network-based query graph ranking, they suffer from three new challenges when handling more complex questions: 1) complicated SPARQL syntax, 2) huge search space, and 3) locally ambiguous query graphs. In this paper, we provide a new solution. As a preparation, we extend the query graph by treating each SPARQL clause as a subgraph consisting of vertices and edges and define a unified graph grammar called AQG to describe the structure of query graphs. Based on these concepts, we propose a novel end-to-end model that performs hierarchical autoregressive decoding to generate query graphs. The high-level decoding generates an AQG as a constraint to prune the search space and reduce the locally ambiguous query graph. The bottom-level decoding accomplishes the query graph construction by selecting appropriate instances from the preprepared candidates to fill the slots in the AQG. The experimental results show that our method greatly improves the SOTA performance on complex KGQA benchmarks. Equipped with pre-trained models, the performance of our method is further improved, achieving SOTA for all three datasets used.
\end{abstract}

\begin{IEEEkeywords}
Knowledge Graph, Question Answering, Formal Language, Query Graph
\end{IEEEkeywords}}

\maketitle

\IEEEdisplaynontitleabstractindextext

% \fi
\IEEEpeerreviewmaketitle

\IEEEraisesectionheading{\section{Introduction}\label{sec:introduction}}


\IEEEPARstart{K}{nowledge} graphs (KGs) are receiving increasing attention~\cite{DBLP:conf/acl/BerantL14,DBLP:conf/acl/YihCHG15,DBLP:conf/emnlp/Hu0Z18} as a valuable form of structured data. SPARQL is a machine-readable logical formal language that is used as a standard interface to efficiently access KGs. Unfortunately, SPARQL is not user-friendly, as it requires in-depth knowledge of its syntax and KG. Humans are used to using natural language questions (NLQs) to describe their need for knowledge. Therefore, how to translate NLQs into proper SPARQL queries is a pressing issue for implementing intelligent question answering over KG (KGQA)~\cite{DBLP:journals/kbs/JiaoWZWFW21,DBLP:journals/kbs/XiongWTWL21}.

% There is a natural semantic gap between SPARQL and NLQ because the former aims to facilitate data operation while the latter aims for daily communication. In order to cross the gap, intermediate representations are proposed as equivalent to SPARQL but hide the specific implementations (e.g., \texttt{WHERE}). The upper part of Fig. \ref{fig:challenges}b shows a \textit{query graph}~\cite{DBLP:conf/acl/YihCHG15}, one of the most widely-used intermediate representations. It connects the \textit{topic entity} in the NLQ to the \textit{answer} through a \textit{core relation chain} with \textit{constraint nodes}, which concisely and intuitively reflects the intention of the NLQ.  

\begin{figure*}[!t]
\centering
\includegraphics[width=\textwidth]{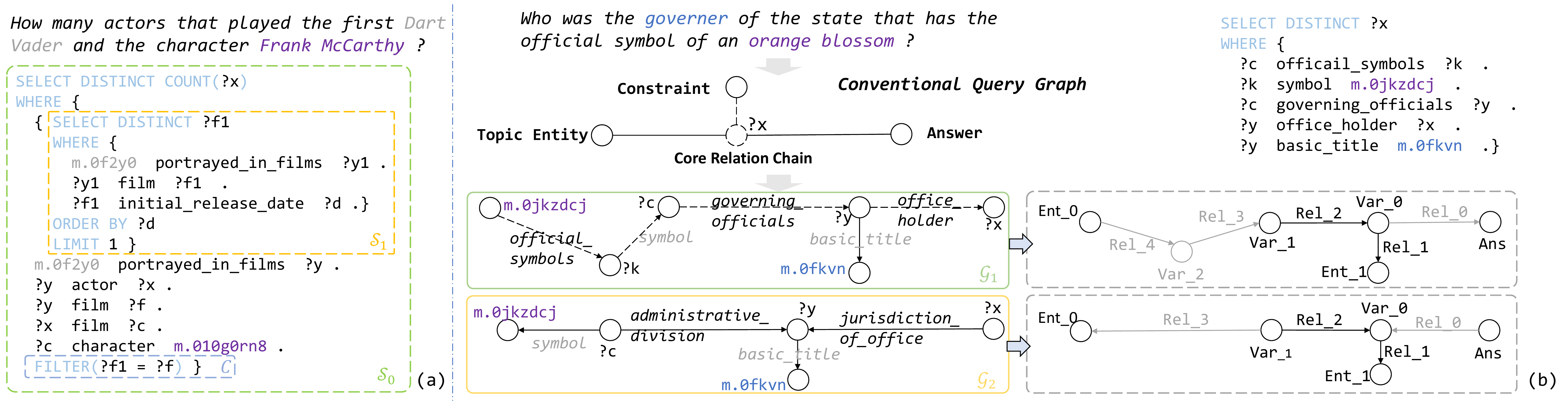}
\caption{Examples of the challenges. (a) depicts a complex SPARQL program with a nested query in ComplexWebQuesions~\cite{DBLP:conf/naacl/TalmorB18}. (b) illustrates a local ambiguity. The green and gold boxes represent the correct and incorrect query graphs, and the dashed boxes represent their structures. For clarity, we only use the abbreviation of each relationship here.}
\label{fig:challenges}
\end{figure*}

There is a natural semantic gap between SPARQL and NLQ, as the former aims to facilitate data manipulation while the latter aims at everyday communication. To bridge this gap, previous work~\cite{DBLP:conf/acl/BerantL14,DBLP:conf/acl/YihCHG15} proposed an intermediate representation comparable to SPARQL but hiding the specific implementation (e.g., \texttt{WHERE}). The upper part of Fig. \ref{fig:challenges}b shows a \textit{query graph}~\cite{DBLP:conf/acl/YihCHG15}, which is one of the most widely used intermediate representations. It connects the \textit{topic entity} in an NLQ to the \textit{answer} through a \textit{core relation chain} with \textit{constraint nodes}, which succinctly and intuitively reflects the intent of the NLQ.  

% In recent work~\cite{DBLP:conf/emnlp/LuoLLZ18,DBLP:conf/semweb/MaheshwariTLCF019,DBLP:conf/wsdm/HeL0ZW21,DBLP:journals/kbs/BakhshiNMR22}, STAGG-Ranking seems to become a popular way for query graph construction. These methods employ neural networks to embed and score the candidate query graphs, and output the top-scored one as the result. To collect the candidate query graphs, they typically adopt a STAGG~\cite{DBLP:conf/acl/YihCHG15} strategy, which can be summarized as the following three steps: 1) First, enumerate all topic entities from the NLQ. 2) Then, enumerate all relation chains starting from the topic entity and having a length of no more than two hops, as the candidate core relation chains. 3) Finally, enumerate all legal constraints and treat them as the nodes attached to the core relation chain. Previous work~\cite{DBLP:conf/emnlp/LuoLLZ18} has demonstrated that STAGG can cover almost all gold SPARQL queries in most KGQA datasets and provide a high-recall query graph candidate set. Therefore, these methods achieve good results conventional benchmarks~\cite{DBLP:conf/emnlp/BerantCFL13,DBLP:conf/acl/YihRMCS16,DBLP:conf/semweb/TrivediMDL17}.

In recent work~\cite{DBLP:conf/emnlp/LuoLLZ18,DBLP:conf/semweb/MaheshwariTLCF019,DBLP:conf/wsdm/HeL0ZW21,DBLP:journals/kbs/BakhshiNMR22}, Learn-to-Rank (LR) seems to be a popular method of query graph construction. It employs a neural network to embed and score the candidate query graphs and outputs the highest scoring graph as the result. To collect candidate query graphs, the LR-based methods usually adopt the STAGG~\cite{DBLP:conf/acl/YihCHG15} strategy, which can be summarized in the following three steps. 1) First, enumerate all topic entities from the NLQ. 2) Then, all relation chains that start from topic entities and are no longer than two hops are enumerated as candidate core relation chains. 3) Finally, all legal constraints are enumerated and treated as nodes of core relation chains. Existing research~\cite{DBLP:conf/emnlp/LuoLLZ18} has shown that STAGG can cover almost all golden SPARQL queries in most KGQA datasets and provide a high recall set of query graph candidates. As a result, these LR-based methods achieve good results on conventional benchmarks~\cite{DBLP:conf/emnlp/BerantCFL13,DBLP:conf/acl/YihRMCS16}.

\subsection{Challenge}
% \noindent \textbf{Challenge}
However, when dealing with some newly released and more difficult datasets such as ComplexWebQuesitions (CWQ)~\cite{DBLP:conf/naacl/TalmorB18}, LR suffers from the following three challenges.

% 1) \textit{Unrepresentable SPARQL Syntax.} The complicated SPARQL query often comprises multiple \texttt{FILTER} clauses, aggregation functions, and even nested queries. For example, Fig \ref{fig:challenges}a shows a gold SPARQL query $\mathcal{S}_0$ (green box) in CWQ. Unlike the queries in traditional datasets, it contains a nested sub-query $\mathcal{S}_1$ (golden box). The existing query graph grammar can not represent this complex syntax with a single constraint node, thus STAGG cannot cover these SPARQL queries.

1) \textit{Non-Representable SPARQL syntax.} Complex SPARQL queries often include multiple \texttt{FILTER} clauses, aggregate functions, and even nested queries. For example, Fig. \ref{fig:challenges}a shows a gold SPARQL query $\mathcal{S}_0$ (green box) in CWQ. Unlike the query in a traditional dataset, it contains a nested subquery $\mathcal{S}_1$ (golden box). The existing query graph syntax is unable to represent such nested relationships with a single constraint node, resulting in STAGG being unable to cover such complicated SPARQL queries.

% 2) \textit{Unbearable Search Space.} Since STAGG needs to enumerate all relation chains that do not exceed $L$ hops, it can obtain approximate $\mathcal{Y} \approx Y^L$ candidate query graphs, where $Y$ is the average number of one-hop relations. This exponential relationship results in a exploded search space (hundreds of thousands) when L is greater than 2 (common in CWQ). Considering the time and space limitations of the devices, such a huge size makes it impractical to train a neural network to embed and rank all the candidate graphs.

2) \textit{Unbearable search space.} Since STAGG needs to enumerate all relation chains that do not exceed $L$ hops, it can obtain approximately $\mathcal{Y} \approx Y^L$ candidate query graphs, where $Y$ is the average number of one-hop relations. When L is greater than 2 (common in CWQ), this exponential relationship leads to an explosive search space, e.g. hundreds of thousands of candidate query graphs. Such a large size makes 
LR impractical to train a neural network to embed and rank all the candidate graphs, considering the time and space limitations of the device.

% 3) \textit{Local Ambiguity.} Even if the search space is of suitable size, STAGG also produces some confusing candidate query graphs because of its enumeration. These graphs do not reflect the true semantics of the NLQ but have some components locally correct. For example, in Fig. \ref{fig:challenges}b, $\mathcal{G}_1$ (blue box) is the correct query graph and $\mathcal{G}_2$ (red box) is a confusing one while has the correct entities \texttt{m.0jkzdcj} and \texttt{m.0fkvn}, and correct relations \texttt{symbol} and \texttt{basic\_title}. We find that the existing ranking model often suffers from these local ambiguous components thence making the wrong predictions.

3) \textit{Local ambiguity.} Even if the size of the search space is appropriate, STAGG produces some confusing candidate query graphs owing to its enumeration. These graphs do not reflect the true meaning of the NLQ, but have some components that are locally correct. For example, Fig. \ref{fig:challenges}b shows a correct query graph $\mathcal{G}_1$ (green box)and an incorrect query graph $\mathcal{G}_2$ (golden box). Although $\mathcal{G}_2$ ultimately fails to query the ``governor" correctly, it has the correct entities \texttt{m.0jkzdcj} and \texttt{m.0fkvn}, and the correct relations \texttt{symbol} and \texttt{basic\_title}. We find that existing LR-based models are frequently influenced by these locally ambiguous components and thus make erroneous predictions.

\subsection{Motivation}
% \noindent \textbf{Motivation}
In this paper, we focus on how to address these three challenges.

For challenge 1, we notice that regardless of how complex the SPARQL syntax is, it can always be decomposed into entities, variables, values, relations, and other built-in properties. If these fine-grained components are treated as vertices and edges, the resulting query graph should be more capable of representation.

For challenge 2, some related work~\cite{DBLP:conf/emnlp/YuYYZWLR18,DBLP:conf/acl/GuoZGXLLZ19,DBLP:conf/acl/WangSLPR20} reveals that \textit{generation} can decompose the large total search space into smaller subspaces. If the query graph is generated in multiple steps, the model only needs to embed candidate actions in each step, avoiding the prohibitively expensive time and space cost of embedding all candidate graphs.

For challenge 3, we hypothesize that the structure of the query graph is effective at disambiguation. Consider the example in Fig. \ref{fig:challenges}b. The topologies of $\mathcal{G}_1$ and $\mathcal{G}_2$ are depicted by the two dashed boxes. Although the components of these two query graphs are similar locally, their structures are vastly different. In comparison to $\mathcal{G}_1$, $\mathcal{G}_2$ has one less variable node, \texttt{Var\_2}, and the relations \texttt{Rel\_0} and \texttt{Rel\_3} point in the wrong direction.
We analyzed the bad cases caused by local ambiguity on LC-QuAD and found that 69\% of them had incorrect structures. Hence, we give a reasonable hypothesis that the effect of local ambiguity can be mitigated if the structure is correctly predicted in the first place. In fact, our hypothesis has a more fundamental motivation: the search space for structure prediction is much smaller than for instance prediction. Even the most difficult CWQ has only a few dozen different query graph structures but contains more than 6,000 different relations (instances). In general, the smaller the search space is, the easier the task.
Thus, structure prediction is a simpler and more critical task that deserves to be solved first.

\begin{figure*}[!t]
\centering
\includegraphics[width=\textwidth]{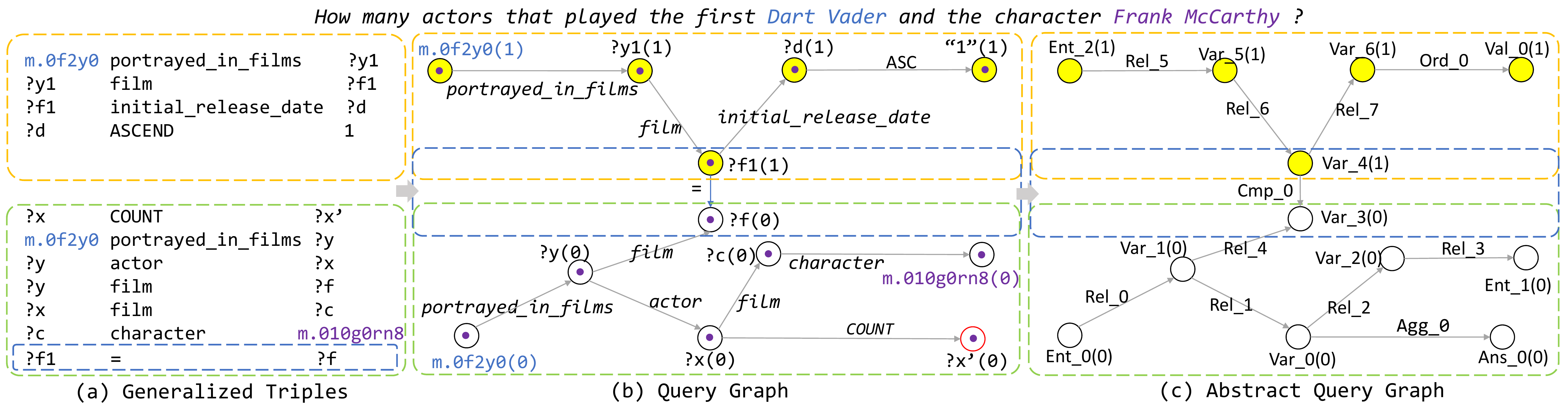}
\caption{Examples of generalized triples, query graphs and abstract query graphs. Two dashed boxes of the same color indicate that they correspond. In (b) and (c), the number in parentheses after each vertex name indicates its segment number. The red circle, \texttt{?x'(0)} in (b), is the vertex representing the answer.}
\label{fig:query_graph}
\end{figure*}

\subsection{Our Method}
% \noindent \textbf{Our method}
We first redefine a fine-grained query graph by treating each clause of a SPARQL query as a subgraph composed of vertices and edges to represent the complicated syntax. Subsequently, to reduce the search space and avoid local ambiguity, we propose an abstract query graph (AQG) to describe the structure of our query graph. It preserves the topology of the query graph, but replaces each instance with a slot of categories (e.g., entities, relations, and values), thereby acting as a structural constraint.

Based on the proposed AQG grammar, we propose a \textit{Hierarchical Graph Generation Network} (HGNet) to generate query graphs, which consists of two stages.
In the first stage, for each category, the top $k$ relevant instances are collected as a candidate instance pool by a simple strategy.
In the second stage, HGNet first encodes the NLQ and then performs autoregressive decoding to generate the query graph from scratch.
In contrast to previous work~\cite{DBLP:conf/acl/LanJ20,DBLP:conf/emnlp/QinLPA21,DBLP:journals/information/LiuTGL21}, our decoding procedure is hierarchical and consists of two phases, \textit{Outlining} and \textit{Filling}. \textit{Outlining} starts with an empty graph and aims to generate an AQG. At each decoding step, the model extends the graph by predicting and adding a vertex/edge slot until it stops by itself. \textit{Filling} begins with a completed AQG and proceeds to generate a query graph. At each decoding step, the model predicts an instance from the candidate instance pool and populates it with the corresponding slot vertices/edges. The query graph is completed when all the slots are filled. Unlike the common Seq2Seq model, HGNet uses a graph neural network to obtain a vector representation of the AQG for each step during \textit{Outlining}, forcing itself to always be aware of the structural information of the AQG.
We conducted comprehensive experiments on three benchmarks. Our proposed HGNet achieves a significant improvement (20.9\%) on the most challenging CWQ~\cite{DBLP:conf/naacl/TalmorB18} and competitive results on LC-QuAD~\cite{DBLP:conf/semweb/TrivediMDL17} and WebQSP~\cite{DBLP:conf/acl/YihRMCS16}.  With the support of pretrained \textsc{Bert}~\cite{DBLP:conf/naacl/DevlinCLT19}, HGNet outperforms all the existing methods on all three datasets.

Overall, the contributions of this paper can be summarized as follows:

1) We extend the existing definition of query graphs to accommodate the complicated SPARQL syntax and propose a fine-grained AQG grammar to describe the structure of query graphs. This AQG normalizes the classes of vertices and edges to provide prerequisites for grammar-based end-to-end generation.

2) We propose an end-to-end HGNet model that follows a hierarchical query graph generation framework. It leverages AQG as a structural constraint to narrow the search space and reduce the local ambiguity of the query graph. To the best of our knowledge, this is the first time that a query structure has been utilized as an explicit constraint for end-to-end query graph generation.

3) We conducted comprehensive experiments on three commonly used datasets and demonstrated that HGNet has a significant advantage in solving complex NLQs. With the support of pre-trained models, HGNet outperforms all compared methods.

\section{Preliminaries}

\subsection{Knowledge Graph}

A knowledge graph (KG) is typically a collection of subject-predicate-object triples, denoted by $\mathcal{K}=\{\left \langle s, p, o\right \rangle | s \in \mathcal{E}, p \in \mathcal{R}, o \in \mathcal{E} \cup \mathcal{L} \}$, where $\mathcal{E}$, $\mathcal{R}$ and $\mathcal{L}$ denote the entity set , relation set, and literal set, respectively. If $o$ is an entity, $\left \langle s, p, o\right \rangle$ denotes that relation $p$ exists between head entity $s$ and tail entity $o$. If $o$ is a literal, $\left \langle s, p, o\right \rangle$ denotes that entity $s$ has property $p$, the value of which is $o$. 
%Here, literal can be a date (e.g., 2015-08-15), an entity alias (e.g., \textit{Dart Vader}), or a numeric value (e.g., "1").

\subsection{Query Graph}
\label{sec:query_graph}
We first redefine the query graph as follows.

\begin{myDef}[Query Graph]
A query graph is a directed acyclic graph, denoted by $\mathcal{G}_q = (V_q, E_q, \Psi_q, \Phi_q, S_q)$. Here, $V_q$ is the vertex set and $E_q =\{e|e=\left \langle v, v'\right \rangle, v,v' \in V_q\}$ is the edge set. $\Psi_q = \{(v,l_v)|v \in V_q\}$, where $l_v \in \Psi$ is a vertex instance and $\Psi$ denotes the set of entities, types, values, and variables. $\Phi_q = \{(e, l_e)| e \in E_q\}$, where $l_e \in \Phi$ is an edge instance and $\Phi$ denotes the set of relations and built-in properties. $S_q = \{(v, s_v)|v \in V_q\}$, where $s_v$ denotes the segment number of $v$.
\end{myDef}

% The redefined query graph is able to represent \texttt{FILTER}, \texttt{ORDER BY}, aggregation functions, and nested queries. Fig. \ref{fig:query_graph}a illustrates how to convert the complicated SPARQL in our runing example to a query graph. First, the SPARQL query is parsed to generalized triples, of which the predicate can be either a KG relation or built-in property, such as relation triples (e.g.,$\left \langle \texttt{?y1},\texttt{film}, \texttt{?f1} \right \rangle$), comparison triples (e.g.,$\left \langle \texttt{?f1},\texttt{=}, \texttt{?f} \right \rangle$), order triples (e.g.,$\left \langle \texttt{?d},\texttt{ASCEND}, \texttt{1} \right \rangle$), and aggregation triples (e.g., $\left \langle \texttt{?x},\texttt{COUNT}, \texttt{?x'} \right \rangle$). All the triples are extracted from the main query (green box) and each sub-query (golden box) and then merged. For each triple, the subject and object are regarded as vertices, and the predicate as an edge, then the query graph (Fig. \ref{fig:query_graph}b) is obtained. Appendix \ref{app:pre-process_sparql} details the solutions to more complicated cases. The query graph has the following two properties.
The redefined query graph can represent \texttt{FILTER}, \texttt{ORDER BY}, aggregate functions and nested queries. Fig. \ref{fig:query_graph}a illustrates how to convert the complex SPARQL of our running example into a query graph. First, SPARQL queries are parsed into generalized triples where the predicates can be KG relations or built-in properties, such as relation triples (e.g.,$\left \langle \texttt{?y1},\texttt{film}, \texttt{?f1} \right \rangle$), comparison triples (e.g.,$\left \langle \texttt{?f1},\texttt{=}, \texttt{?f} \right \rangle$), ordinal triples (e.g.,$\left \langle \texttt{?d},\texttt{ASCEND}, \texttt{1} \right \rangle$), and aggregation triples (e.g., $\left \langle \texttt{?x},\texttt{COUNT}, \texttt{?x'} \right \rangle$). All triples are extracted from the main query (green box) and each subquery (gold box) and then merged. For each triple, the subject and object are considered as vertices and the predicate as an edge, and then the query graph is obtained (Fig. \ref{fig:query_graph}b). Appendix \ref{app:pre-process_sparql} details the solution for more complex cases. We specify that any query graph has the following two properties.
\begin{myProp}
$\forall v_1,\forall v_2 \in V_q$, 
%$(v_1, s_{v_1}), (v_2, s_{v_2}) \in S_q$, 
if $s_{v_1} = s_{v_2}$, then $v_1$ and $v_2$ are from the same segment of the SPARQL.
\end{myProp}

We use a \textit{segment} to represent a subquery or main query of SPARQL. In our query graph, the segment number of the main query is always set to 0, as in the green box of Fig. \ref{fig:query_graph}b.
\begin{myProp}
\label{prop:vertex_number}
$\forall \mathcal{G}_q = (V_q, E_q, \Psi_q, \Phi_q, S_q)$, $|V_q| = |E_q| + 1$.
\end{myProp}
% Obviously, a segment is weakly connected and any meaningful sub-segment ($s_v > 0$) is associated with the main segment ($s_v = 0$) through the \texttt{FILTER} clause (red box). Therefore, a query graph is always weakly connected, i.e. $|V_q| \le |E_q| + 1$. This paper further simplifies the problem by focusing on the query graphs that satisfy $|V_q| = |E_q| + 1$. Although the assumption is strengthened, in the hardest CWQ~\cite{DBLP:conf/naacl/TalmorB18} dataset, there are still 98.5\% of cases meet it. 

A segment is clearly a weakly connected graph, and any meaningful subsegment ($s_v > 0$) is associated with the main segment ($s_v = 0$) via the \texttt{FILTER} clause (blue box in Fig. \ref{fig:query_graph}b). Consequently, a query graph is always weakly connected, i.e., $|V_q| \le |E_q| + 1$. This paper further simplifies the problem by focusing on query graphs that satisfy $|V_q| = |E_q| + 1$. Although the hypothesis is strengthened, it is still satisfied in 98.5\% of the cases in the CWQ~\cite{DBLP:conf/naacl/TalmorB18} dataset.

Here our redefinition induces multiple \texttt{FILTER} clauses, \texttt{ORDER BY} clauses, aggregation functions and nested queries into a unified graph grammar in preparation for query graphs that can be generated using grammar-based autoregressive decoding. In the rest of this paper, all query graphs refer to our redefined query graphs.

\subsection{Abstract Query Graph}
\label{sec:aqg}

Considering that the structure of a query graph is reflected in the topology and categories of its vertices and edges, we define an abstract query graph (AQG).

\begin{myDef}[Abstract Query Graph]
An abstract query graph is a directed acyclic graph, denoted by $\mathcal{G}_a=(V_a, E_a, \Psi_a, \Phi_a, S_a)$. Here, $V_a$ is the set of vertices and $E_a =\{e|e=\left \langle v, v'\right \rangle, v,v' \in V_a\}$ is the set of edges. $\Psi_a = \{(v, c_v)|v \in V_a\}$, where $c_v \in \mathbf{\Psi}$ is a class label and $\mathbf{\Psi}=$ \{\texttt{Ans}, \texttt{Var}, \texttt{Ent}, \texttt{Type}, \texttt{Val}\}. $\Phi_a = \{(e, c_e)|e \in E_q\}$, where $c_e \in \mathbf{\Phi}$ is a class label and $\mathbf{\Phi}= $ \{\texttt{Rel}, \texttt{Ord}, \texttt{Cmp}, \texttt{Agg}\}. $S_a = \{(v, s_v)|v \in V_a\}$, where $s_v$ denotes the segment number of $v$. 
\end{myDef}

Note that the key difference between the AQG and the query graph is that all the vertices and edges of the AQG are class labels, not real instances.
\texttt{Ans} is the class of answers, i.e., \texttt{?x} in Fig. \ref{fig:query_graph}b; \texttt{Var} is the class of variables, i.e., placeholders of variables other than answers, such as \texttt{?y} and \texttt{?f1}. \texttt{Ent} indicates the class of entities, e.g., \texttt{m.0f2y0}; \texttt{Type} represents the class of entity types, e.g., \texttt{Actor} (in \textit{DBPedia}) and \texttt{award.award\_winner} (in \textit{Freebase}); \texttt{Val} denotes the class of values, including \texttt{"1"} and \texttt{"2011-01-01"}; \texttt{Rel} is the class of KG relations, e.g., \texttt{portrayed\_in\_films}; and \texttt{Ord} indicates the class of \{\texttt{ASC}, \texttt{DESC}\} in \texttt{ORDER BY} clauses. For example, triple $\left \langle \texttt{?v ASC 3} \right \rangle$ means \texttt{ORDER BY ASC(?v) LIMIT 3}; \texttt{Cmp} denotes the class of \{$=$, $\ne$, $>$, $\ge$, $<$, $\le$, \texttt{DURING}, \texttt{OVERLAP}\} that describes the comparison relationships between variables and values. Here \texttt{DURING} and \texttt{OVERLAP} are used to handle time intervals, see Appendix \ref{app:temporal_constraint} for details; \texttt{Agg} represents the class of aggregation functions \{\texttt{COUNT}, \texttt{MAX}, \texttt{MIN}, \texttt{ASK}\}. For example, triple $\left \langle \texttt{?x COUNT ?y} \right \rangle$ means that \texttt{?y} is the count number of variable \texttt{?x}. 

Fig. \ref{fig:query_graph}c exhibits the AQG of the running example. Since the AQG inherits the topology and segments of the query graph, it naturally has the following properties.
\begin{myProp}
$\forall v_1,\forall v_2 \in V_a$, $(v_1, s_{v_1}), (v_2, s_{v_2}) \in S_a$, if $s_{v_1} = s_{v_2}$, then $v_1$ and $v_2$ correspond to the same segment.
\end{myProp}
\begin{myProp}
$\forall \mathcal{G}_a = (V_a, E_a, \Psi_a, \Phi_a, S_a)$, $|V_a| = |E_a| + 1$.
\label{prop:v_number}
\end{myProp}
Each vertex and edge of the AQG can be regarded as a slot of the abstract class. A query graph $\mathcal{G}_q$ can be converted to a unique AQG $\mathcal{G}_a$ by replacing the instances with the corresponding class slots; in contrast, an AQG $\mathcal{G}_a$ can also generate a series of query graphs $\{\mathcal{G}_q^i\}$ by filling all the class slots with a combination of instances. Formally, let $\mathcal{F}: \mathbf{G_Q} \rightarrow \mathbf{G_A}$ be an $n$-to-one mapping, where $\mathbf{G_Q}$ and $\mathbf{G_A}$  are the domains of query graphs and AQGs, respectively, and $\mathcal{F}^{-1}: \mathbf{G_A} \rightarrow \mathbf{G_Q}$ is the inverse one-to-$n$ mapping of $\mathcal{F}$; then, there are the following two properties.
\begin{myProp}
$\forall \mathcal{G}_q^1,\forall \mathcal{G}_q^2 \in \mathbf{G_Q}$ is structurally equivalent, if and only if $\mathcal{F}(\mathcal{G}_q^1) = \mathcal{F}(\mathcal{G}_q^2)$.
\end{myProp}
\begin{myProp}
\label{prop:disjoint}
$\forall \mathcal{G}_a^1,\forall \mathcal{G}_a^2 \in \mathbf{G_A}$, if $\mathcal{G}_a^1 \neq \mathcal{G}_a^2$, then $\mathcal{F}^{-1}(\mathcal{G}_a^1) \cap \mathcal{F}^{-1}(\mathcal{G}_a^2) = \emptyset$.
\end{myProp}
Property \ref{prop:disjoint} reveals that two resulting query graph sets from different AQGs are disjoint. As a result, the search space of $\mathcal{G}_q$ can be significantly reduced from the full domain $\mathbf{G_Q}$ to $\mathbf{G'_Q}$ if AQG $\mathcal{G}_a$ is determined first, where $\mathbf{G'_Q}=\mathcal{F}^{-1}(\mathcal{G}_a)$ and $|\mathbf{G'_Q}| \ll |\mathbf{G_Q}|$.

Previous work has proposed some concepts of query graph structure, such as SQG~\cite{DBLP:conf/emnlp/Hu0Z18} and \textit{query structure}~\cite{DBLP:conf/emnlp/DingHXQ19}. However, the former consists of NLQ phrases, and the latter ignores the categories of built-in properties and cannot handle nested structures. In contrast, our AQG focuses on query graphs and normalizes the categories of vertices and edges for the first time, which can describe complex structures including nested queries.

\begin{figure}[!t]
\centering
\includegraphics[width=0.48\textwidth]{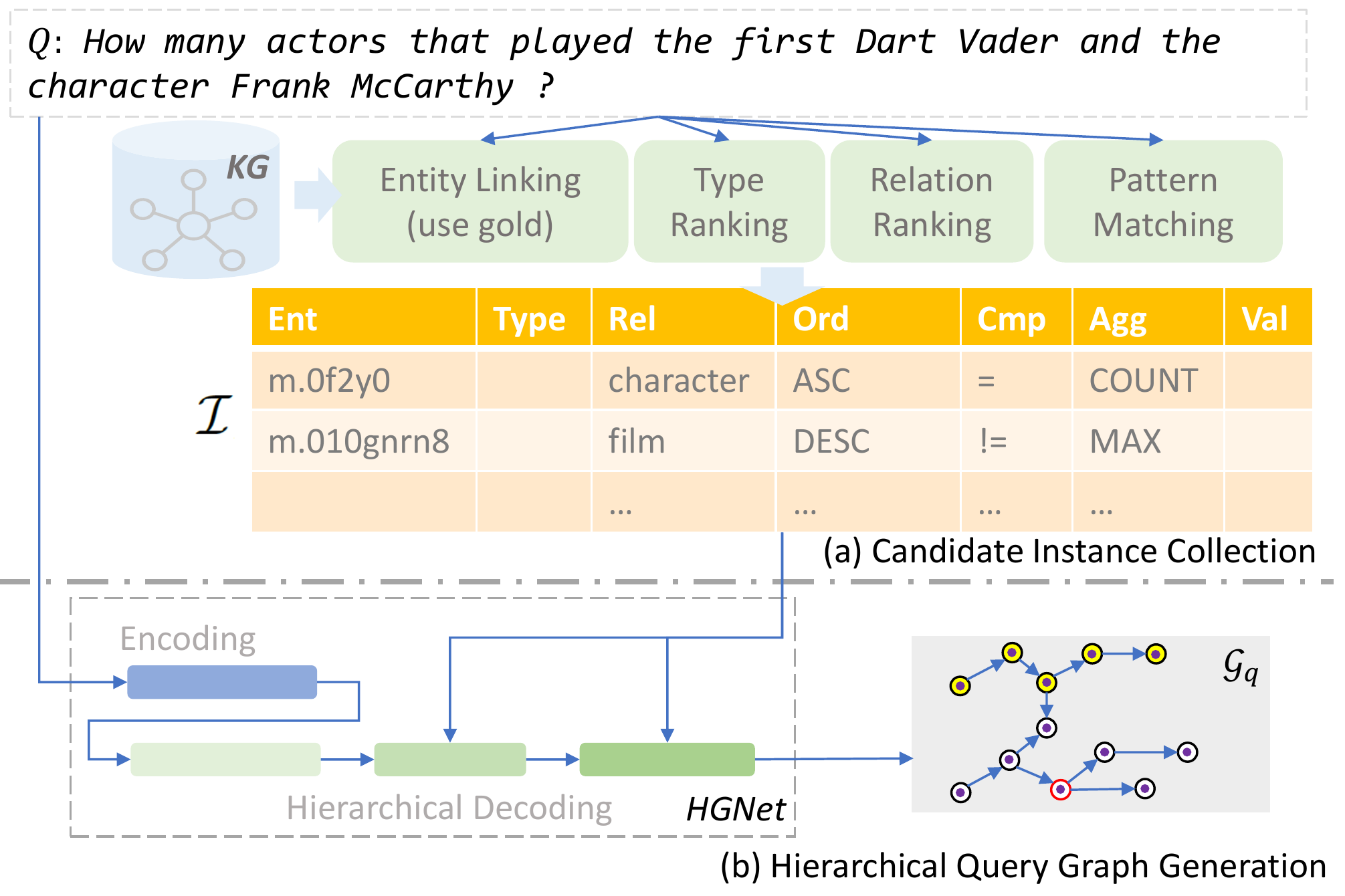}
\caption{Entire process of our proposed method.}
\label{fig:overview}
\end{figure}

\section{Overview}
Fig. \ref{fig:overview} shows an overview of our proposed method. Given a natural language question $\mathcal{Q}$ and a knowledge graph $\mathcal{K}$, the target query graph $\mathcal{G}_q$ is constructed by the following two stages.

% \subsection{Candidate Instance Collection}
\noindent \textbf{Candidate Instance Collection.} The construction of the query graph requires specific instances of entities, relationships, etc. Considering the efficiency, it is not practical to use all instances in the KG as candidate instances every time. Hence, this stage obtains a pool of candidate instances for HGNet with high recall. As mentioned in Section \ref{sec:aqg}, the classes of instances can be generalized as $\mathbf{\Psi} \cup \mathbf{\Phi} = $ \{\texttt{Ans}, \texttt{Var}, \texttt{Ent}, \texttt{Rel}, \texttt{Type}, \texttt{Val}, \texttt{Ord}, \texttt{Cmp}, \texttt{Agg}\}. Except for the noninstance classes \texttt{Ans} and \texttt{Var}, the candidate instances of the remaining classes are collected by type linking, relationship linking and pattern matching, respectively, as detailed in Section \ref{sec:stage1}.
The obtained pool is denoted by $\mathcal{I} = \{\mathcal{I}_c | c \in \mathbf{\Psi} \cup \mathbf{\Phi} - \{\texttt{Ans}, \texttt{Var}\} \} $, where $\mathcal{I}_c$ is the set of candidate instances of class $c$. 

% \subsection{Hierarchical Query Graph Generation}
\label{sec:overview_stage2}

\noindent \textbf{Hierarchical Query Graph Generation.} Having $\mathcal{I}$, our HGNet generates a query graph $\mathcal{G}_q$ by autoregressive decoding of the following two levels.

% At the high-level decoding process, HGNet starts from an empty graph and expands it to an AQG $\mathcal{G}_a$ by a sequence of the graph-based operations. Specifically, at each step, the model first predicts an operation with the combined information of the question and previous graph and then executes it to obtain a new graph. We call this process \textit{Outlining}, detailed in \ref{sec:outlining}.
In the high-level decoding process, HGNet starts with an empty graph and expands it to AQG $\mathcal{G}_a$ through a series of graph-based operations. Specifically, at each step, the model first predicts an operation using the combined information of the NLQ and the graph obtained in the previous step and then performs that operation to obtain a new graph. We refer to this process as textit{Outlining}; see \ref{sec:outlining} for details.

% At the ground-level decoding process, HGNet begins with $\mathcal{G}_a$ and instantiates it to a query graph $\mathcal{G}_q$ by a sequence of operations. Concretely, at each step, the model focuses on a slot in $\mathcal{G}_a$, denoted by $x$, and predicts an instance $i \in \mathcal{I}_{c_x}$ to fill $x$. We call this process \textit{Filling}, detailed in \ref{sec:filling}. When all the slots (except \texttt{Var} and \texttt{Ans}) in $\mathcal{G}_a$ are filled with the instances, final query graph $\mathcal{G}_q$  is completed.
In the bottom decoding process, HGNet starts from $\mathcal{G}_a$ and instantiates $\mathcal{G}_a$ into a query graph $\mathcal{G}_q$ through a series of fill operations. Concretely, at each step, the model focuses on a slot in $\mathcal{G}_a$, denoted by $x$, and predicts an instance $i$ from $\mathcal{I}_{c_x}$ to fill $x$. We call this process \textit{Filling}; see \ref{sec:filling} for details. $\mathcal{G}_q$ is completed when all the slots in $\mathcal{G}_a$ (except \texttt{Var} and \texttt{Ans}) are filled with instances.

\section{Candidate Instance Collection}
\label{sec:stage1}

For classes \texttt{Ord}, \texttt{Cmp}, and \texttt{Agg}, candidate instances are obtained by enumeration. Specifically, $\mathcal{I}_{ord} = $ \{\texttt{ASC}, \texttt{DESC}\}, $\mathcal{I}_{cmp} = $ \{$=$, $\ne$, $>$, $\ge$, $<$, $\le$, \texttt{DURING}, \texttt{OVERLAP}\}, and $\mathcal{I}_{agg} = $ \{\texttt{COUNT}, \texttt{MAX}, \texttt{MIN}, \texttt{ASK}\}. These built-in properties are very few, so there is no need to filter them.

For class \texttt{Val}, the set of candidate instances $\mathcal{I}_{val}$ is extracted from NLQ $\mathcal{Q}$ by pattern matching. We summarize the types of values: integers, floating-point numbers, quoted strings, years, and dates. Since these values are often based on specific patterns, we design regular expressions to extract them.

% For class \texttt{Rel}, $\mathcal{I}_{rel}$ is obtained by relation ranking~\cite{DBLP:conf/acl/YuYHSXZ17}. Concretely, we train a naive BiLSTM as the ranker to encode $\mathcal{Q}$ and each one-hop relation $r \in \mathcal{R}$ and calculate their semantic relevance score. The top-$k$ relations with highest score are returned. The ranker is trained by optimizing the hinge loss of positive and negative relations. Here, the positive relations are obtained from the gold SPARQL and the negative are randomly sample from $\mathcal{R}$.
For class \texttt{Rel}, $\mathcal{I}_{rel}$ is obtained by relation ranking ~\cite{DBLP:conf/acl/YuYHSXZ17}. Specifically, we train a naive bidirectional long short-term memory network (BiLSTM) as a ranker to encode $\mathcal{Q}$ and each single-hop relation $r \in \mathcal{R}$, and compute their semantic relevance scores. The $k$ relations with the highest scores are returned. The ranker is trained by optimizing the hinge loss of positive and negative relations. Here, the positive relations are obtained from the gold SPARQL and the negative relations are randomly selected from $\mathcal{R}$, where $\mathcal{R}$ is the relation set of KG $\mathcal{K}$.

For class \texttt{Type}, $\mathcal{I}_{type}$ is obtained by a similar method to \texttt{Rel}. The only difference is that we add a type \texttt{NONE} here to indicate the case where none of the types are in $\mathcal{Q}$. In this way, if the top-1 type is \texttt{NONE}, there is no candidate instance returned.

For class \texttt{Ent}, to make a fair comparison with previous work~\cite{DBLP:conf/emnlp/SunDZMSC18,DBLP:conf/emnlp/SunBC19,DBLP:conf/semweb/MaheshwariTLCF019,DBLP:conf/wsdm/HeL0ZW21}, we follow them to extract gold entities directly from gold SPARQL queries as candidate instances. In our running example, $\mathcal{I}_{ent} =$ \{\texttt{m.0f2y0}, \texttt{m.010gnrn8}\}, where \texttt{m.0f2y0} and \texttt{m.010gnrn8} are the machine codes of entities {Dart\_Vader} and {Frank\_McCary}, respectively. 

It is important to emphasize that the goal of this phase is merely to have a high recall of the set of candidate instances, allowing the use of any alternative method.

\section{Hierarchical Generation Framework}
\label{sec:framework}
Before introducing HGNet, it is necessary to describe in detail the hierarchical generation framework that HGNet follows. The framework relies on our redefined query graph and AQG grammar, which can generate query graphs of various complex structures. It consists of two phases,  \textit{Outlining} and \textit{Filling}.
\subsection{Outlining Process}
\label{sec:outlining}
\begin{figure}[!t]
\centering
\includegraphics[width=0.45\textwidth]{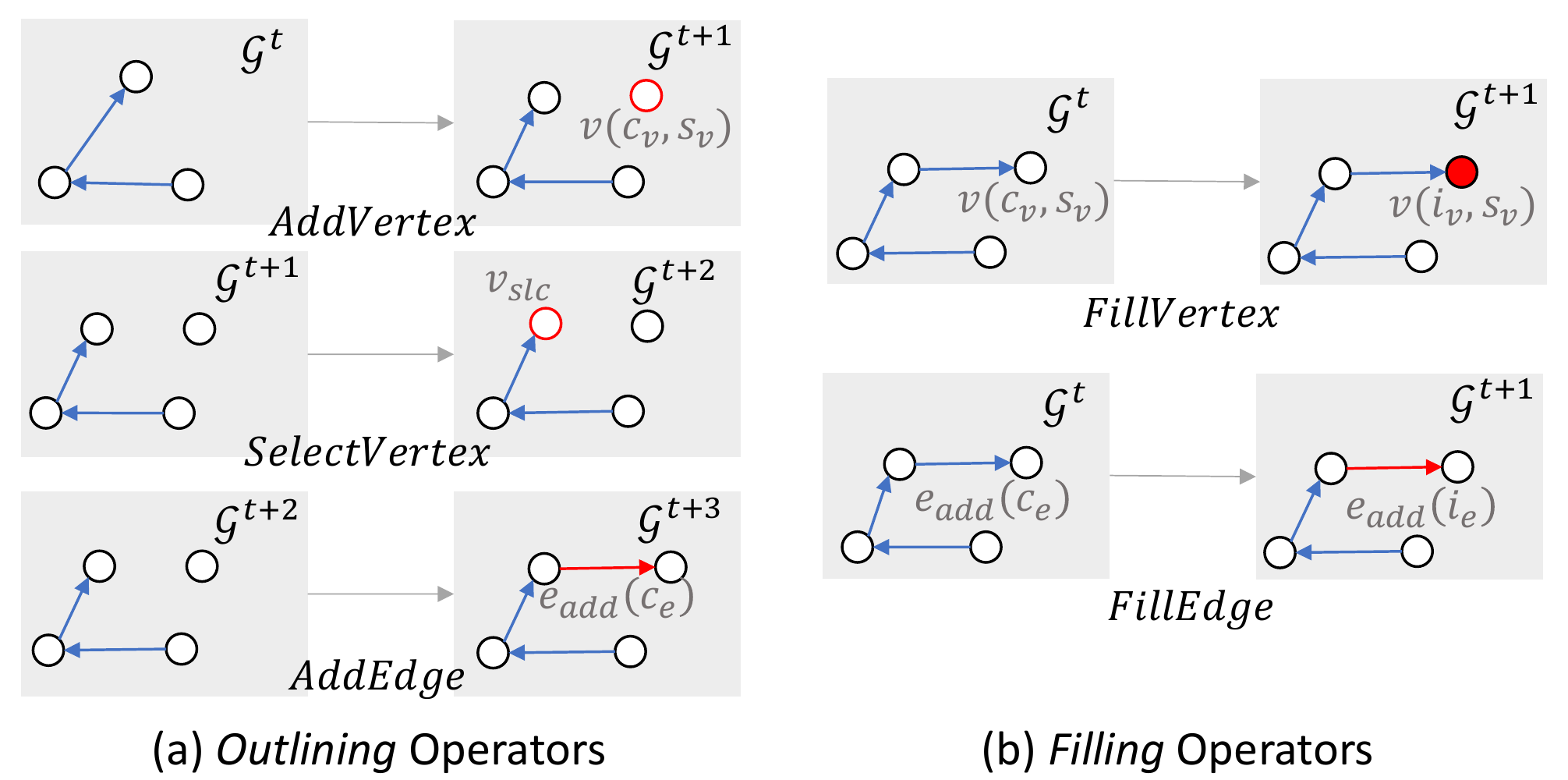}
\caption{Operators in our hierarchical generation framework. The red objects represent the concerned vertices and edges.}
\label{fig:operation}
\end{figure}

The goal of \textit{Outlining} is to generate an AQG with $\mathcal{N}$ vertices and $\mathcal{N}-1$ edges, where $\mathcal{N}$ is unknown. The process can be described by a sequence of graphs, $\{\mathcal{G}^0, \mathcal{G}^1, ..., \mathcal{G}^\mathcal{T}\}$, where $\mathcal{T}=3\mathcal{N}-1$ and $\mathcal{G}^t = (V^t, E^t, \Psi^t, \Phi^t, S^t)$ is the graph generated at step $t$ $(0 \le t \le \mathcal{T})$. In particular, $\mathcal{G}^0 = (\emptyset, \emptyset, \emptyset, \emptyset, \emptyset)$ is an empty graph and $\mathcal{G}^\mathcal{T}$ is the completed AQG, i.e., $\mathcal{G}_a$. When $t \ge 1$, the graph at step $t$ is obtained by performing an 
outlining operation on the graph at step $t-1$, i.e., $\mathcal{G}^t = f^t(\mathcal{G}^{t-1}, *a^t)$. Here, $f^t$ denotes the \textit{Outlining} operator to expand the graph at time step $t$ and $*a^t$ denotes the arguments of $f^t$. According to the characteristics of the AQG, we define the following three \textit{Outlining} operators as shown in Fig. \ref{fig:operation}a. 

    (a) \textit{AddVertex} For graph $\mathcal{G}^t = (V^t, E^t, \Psi^t, \Phi^t, S^t)$, operation \textit{AddVertex}$(\mathcal{G}^t, c_v, s_v, v)$ represents a new vertex $v$ being added in $\mathcal{G}^t$ to obtain a new graph, denoted by $\mathcal{G}^{t+1} = (V^t \cup \{v\}, E^t, \Psi^t \cup \{(v, c_v)\}, \Phi^t, S^t \cup \{(v, s_v)\})$. $c_v \in \mathbf{\Psi} \cup \{\texttt{End}\}$ denotes the class label of $v$; $s_v$ denotes the segment number of $v$. Note that \texttt{End} is an additional class that signals the end. If $c_v=\texttt{End}$, no new vertices will be added in $\mathcal{G}^t$ and \textit{Outlining} terminates. 
    
    (b) \textit{SelectVertex} For graph $\mathcal{G}^{t+1} = (V^t \cup \{v\}, E^t, \Psi^t \cup \{(v, c_v)\}, \Phi^t, S^t \cup \{(v, s_v)\})$, operation \textit{SelectVertex}$(\mathcal{G}^{t+1}, u, v)$ represents  that vertex $u \in V^t$ is selected and will be connected to vertex $v$ in the next step. 
    Note that there is no difference in structure between $\mathcal{G}^{t+1}$ and the new graph returned. However, to explicitly show this operation, the new graph is denoted by $\mathcal{G}^{t+2}.$
    
    (c) \textit{AddEdge} For graph $\mathcal{G}^{t+2} = (V^t \cup \{v\}, E^t, \Psi^t \cup \{(v, c_v)\}, \Phi^t, S^t \cup \{(v, s_v)\})$, \textit{AddEdge}$(\mathcal{G}^{t+2}, c_e, e)$ represents that a new edge $e=\left \langle u,v \right \rangle$ (or $\left \langle u,v \right \rangle$) is added in $\mathcal{G}^{t+2}$. The new graph obtained is denoted by $\mathcal{G}^{t+3} = (V^t \cup \{v\}, E^t \cup \{e\}, \Psi^t \cup \{(v,c_v)\}, \Phi^t \cup \{(e,c_e)\}, S^t \cup \{(v,s_v)\})$, where $c_e \in \mathbf{\Phi}$ is the class label of $e$. Note that the direction of $c_e$ is $+$ or $-$, e.g., \textit{Rel}$+$ means that the relation is from $u$ to $v$, \textit{Rel}$-$ means that it is from $v$ to $u$.

As shown in Fig. \ref{fig:operation}a, successive executions of  \textit{AddVertex}, \textit{SelectVertex}, and \textit{AddEdge} add a new edge $\left \langle u,v \right \rangle$ to the graph. In \textit{Outlining}, $f^t$ is predetermined for each step $t$. Concretely, the first operation is specified as \textit{AddVertex}, which converts the initial empty graph $\mathcal{G}^0 = (\emptyset, \emptyset, \emptyset, \emptyset, \emptyset)$ into $\mathcal{G}^1 = (\{v_0\}, \emptyset, \{(v,c_v)\}, \emptyset, \{0\})$. Then, the 
AQG is gradually generated by executing \textit{AddVertex}, \textit{SelectVertex}, and \textit{AddEdge} in a loop. The last operation is always \textit{AddVertex} to select \texttt{End} to break the loop. Formally, for $t\ge 1$,

\begin{equation}
    f^t=\left\{
    \begin{array}{rcl}
    \textit{AddVertex}       &      & {t = 1 \quad \text{or} \quad t\mod 3 = 2}\\
    \textit{SelectVertex}     &      & {t \mod 3 = 0 }\\
    \textit{AddEdge}     &      & {t > 1 \quad \text{and} \quad t \mod 3 = 1}
    \end{array} \right.
\end{equation}

Although all operators are fixed, different arguments $*a^t$ produce  different AQGs. For example, a different $u$ of \textit{SelectVertex} makes the connection of different vertex pairs and thus results in different topologies. This demonstrates the diversity of the query graphs generated by \textit{Outlining}. 
% However, no matter how variable the structure is, the final graph $\mathcal{G}^\mathcal{T}$ is always connected and its vertex number is always one more than its edge number. Therefore, the product of \textit{Outlining} is always a legitimate AQG. 
Note that the final $\mathcal{G}^\mathcal{T}$ is always a connected graph, no matter how the structure changes. When \textit{Outlining} ends, disregarding the last \textit{AddVertex}(\texttt{End}), the number of \textit{AddVertex} is always one more than the number of \textit{AddEdge}. Thus $\mathcal{G}^\mathcal{T}$ has $|V^\mathcal{T}| = |E^\mathcal{T}| + 1$, where $|V^\mathcal{T}|$ and $|E^\mathcal{T}|$ are the set of vertices and the set of edges of $\mathcal{G}^\mathcal{T}$, respectively. According to Property \ref{prop:v_number}, \textit{Outlining} always produces a legal AQG.

Notably, for \textit{AddVertex}, argument $v$ is only an ID of the added vertex; for \textit{SelectVertex}, argument $v$ is consistent with the previous \textit{AddVertex}; for \textit{AddEdge}, $u$ and $v$ are consistent with previous \textit{AddVertex} and \textit{SelectVertex}, respectively. Consequently, at each time step, only $c_v$ and $s_v$ of \textit{AddVertex}, $u$ of \textit{SelectVertex}, and $c_e$ of \textit{AddEdge} need to be determined.

\begin{figure*}[t]
\centering
\includegraphics[width=\textwidth]{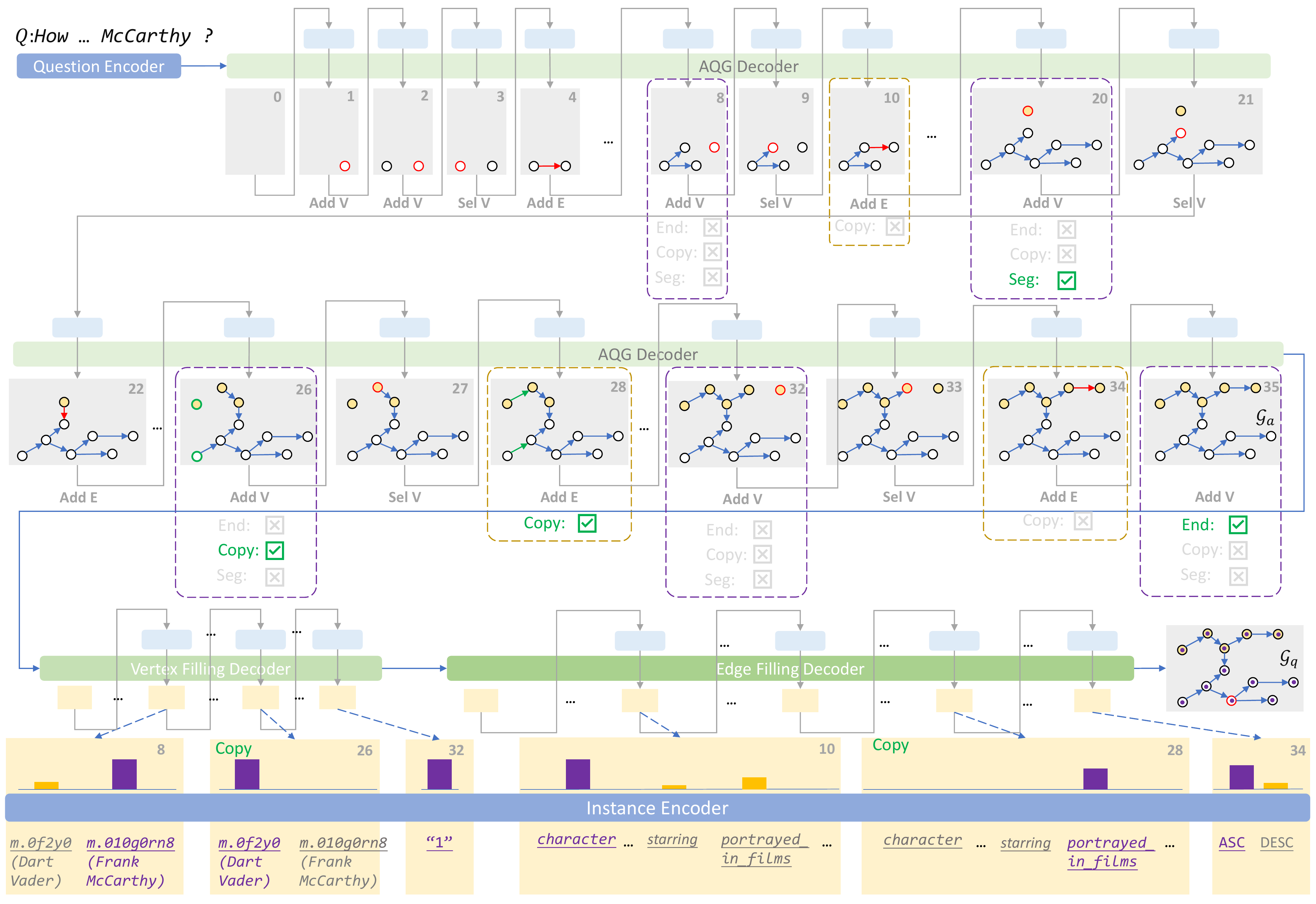}
\caption{The architecture of our proposed HGNet. The gray boxes show the AQG for each step in \textit{Outlining}. The yellow boxes show the bars of the predicted probability for each instance in \textit{Filling}. The dashed boxes indicate some of the steps we are concerned about.}
\label{fig:model}
\end{figure*}

% Compared with STAGG, our graph generation framework greatly increases the coverage of query graph structures. In addition, the framework cooperates with our redefined query graphs to blur the boundaries between entities, relations, built-in properties, etc, which makes it possible to use simple and unified operators instead of a complicated pipeline.

\subsection{Filling Process}
\label{sec:filling}
The goal of \textit{Filling} is to convert the AQG obtained by \textit{Outlining} into a real query graph, whose vertices and edges are all instances rather than class labels. Similar to \textit{Outlining}, \textit{Filling} can also be described by a sequence of graphs, $\{\mathcal{G}^0, \mathcal{G}^1, ..., \mathcal{G}^{\mathcal{T}'}\}$, where $\mathcal{T}'=2\mathcal{N}-1$ and $\mathcal{G}^t = (V^t, E^t, \Psi^t, \Phi^t, S^t)$ is the graph generated at time step $t$ $(0 \le t \le \mathcal{T}')$. In particular, $\mathcal{G}^0 = \mathcal{G}_a$ and $\mathcal{G}^{\mathcal{T}'}$ is the completed query graph $\mathcal{G}_q$. At each step $t \ge 1$, $\mathcal{G}^t = g^t(\mathcal{G}^{t-1}, *b^t)$, where $g^t$ denotes the \textit{Filling} operator used to instantiate a vertex/edge and $*b^t$ indicates the arguments. We define the following two \textit{Filling} operators, as shown in Fig. \ref{fig:operation}b.

(a) \textit{FillVertex} For graph $\mathcal{G}^t = (V^t, E^t, \Psi^t, \Phi^t, S^t)$, operation \textit{FillVertex}$(\mathcal{G}^t, v, i_v)$ represents that vertex $v$ in $\mathcal{G}^t$ is filled with $i_v$ to obtain a new graph, i.e., $\mathcal{G}^{t+1} = (V^t, E^t, \Psi^t - \{(v, c_v)\} \cup \{(v, i_v)\}, \Phi^t, S^t)$. The first argument $v \in V^t$ denotes the vertex to be instantiated and $c_v$ is the class label of $v$. The second $i_v \in \mathcal{I}_{c_v}$ denotes an instance of class $c_v$.
    
(b) \textit{FillEdge} For graph $\mathcal{G}^t = (V^t, E^t, \Psi^t, \Phi^t, S^t)$, operation \textit{FillEdge}$(\mathcal{G}^t, e, i_e)$ represents that edge $e$ in $\mathcal{G}^t$ is filled with $i_e$ to return a new graph, i.e., $\mathcal{G}^{t+1} = (V^t, E^t, \Psi^t, \Phi^t - \{(e, c_e)\} \cup \{(e, i_e)\}, S^t)$. The first argument $e \in E^t$ denotes the edge to be instantiated and $c_e$ is the class label of $e$. The second argument $i_e \in \mathcal{I}_{c_e}$ denotes an instance of class $c_e$.

Compared to vertices, edges are often mentioned implicitly in NLQ (e.g., the relation \textit{portrayed\_in\_films} in Fig. \ref{fig:query_graph}), so intuitively they are more difficult to identify. Inspired by this, we first perform vertex \textit{Filling} and then edge \textit{Filling} to reduce error propagation. Formally, assuming that the generated $\mathcal{G}_a$ has $\mathcal{N}$ vertices and $\mathcal{N}-1$ edges, then  for $t\ge 1$, $g^t$ is determined by

\begin{equation}
    g^t=\left\{
    \begin{array}{rcl}
    \textit{FillVertex}       &      & {t \le \mathcal{N}}\\
    \textit{FillEdge}     &      & {\mathcal{N} < t \le 2\mathcal{N} - 1}
    \end{array} \right.
\end{equation}

The first argument of each \textit{Filling} operation is fixed by \textit{Outlining}. Specifically, at step $t$ of \textit{Filling}, argument $v$ of \textit{FillVertex} denotes the vertex added by \textit{AddVertex} in step $t'=3t-4$ of \textit{Outlining}; argument $e$ of \textit{FillEdge} denotes the edge added by \textit{AddEdge} in step $t'=3t-2+\mathcal{N}$ of \textit{Outlining}. Thus, to accomplish \textit{Filling}, only $i_v$ of \textit{FillVertex} and $i_e$ of \textit{FillEdge} need to be predicted.

\section{Hierarchical Graph Generation Model}

% Fig. \ref{fig:model} shows the architecture of our proposed HGNet, which follows the hierarchical generation framework described in Section \ref{sec:framework} to predict the argument of each pre-determined operation. It leverages the AQG as a constraint to narrow the search space and reduce the local ambiguity. As \textit{Outlining} focuses on structural information while \textit{Filling} on instance information, we employ different decoders for the two processes, respectively. Moreover, to make the model aware of the structure of the entire graph at all times during \textit{Outlining}, we add a graph encoder to embed $\mathcal{G}_{t-1}$ at each step $t$. Our proposed HGNet consists of a NLQ encoder, an instance encoder, a graph encoder, an outlining decoder, a vertex filling decoder, and an edge filling decoder.
Fig. \ref{fig:model} shows the architecture of our proposed HGNet, which follows the hierarchical generation framework described in Section \ref{sec:framework} to predict the parameters of each predefined operation. It utilizes the AQG as a constraint to narrow the search space and reduce local ambiguity. Since \textit{Outlining} focuses on structural information and \textit{Filling} focuses on instance information, we employ different decoders for each of these two processes. In addition, to keep the model informed of the entire graph structure during \textit{Outlining}, we add a graph encoder that embeds $\mathcal{G}_{t-1}$ at each step $t$. Our proposed HGNet consists of an NLQ encoder, an instance encoder, a graph encoder, an outlining decoder, a vertex-filling decoder, and an edge-filling decoder.

\subsection{NLQ Encoder \& Instance Encoder}
To capture the semantic information of NLQ $\mathcal{Q}$, we employ a BiLSTM as the NLQ encoder. It converts the word embeddings of the tokenized $\mathcal{Q}$ to semantic vectors $\mathbf{Q} = [\mathbf{q}_1,\mathbf{q}_2,...,\mathbf{q}_\mathcal{M}] \in \mathbb{R}^{\mathcal{M}d}$, where $\mathcal{M}$ denotes the token number of $\mathcal{Q}$ and $\mathbf{q}_k \in \mathbb{R}^d$ is the vector of the $k$-th word.

The instance encoder is also set to a BiLSTM. For $i \in \mathcal{I}_c$, where $c \in \mathbf{\Psi} \cup \mathbf{\Phi} - \{\texttt{Ans}, \texttt{Var}\}$, its real name (e.g., \texttt{portrayed\_in\_films}) is tokenized and embedded and then fed to the instance encoder. The returned semantic vector is finally converted to a vector representation $\mathbf{h}_i \in \mathbb{R}^d$ using max-pooling. Here, to guarantee that $\mathcal{Q}$ and $\mathcal{I}$ belong to the same semantic space, the parameters of the question encoder and the instance encoder are shared.

\subsection{Graph Encoder}
To keep the model focused on the structural features of the graph during \textit{Outlining}, we apply a graph transformer~\cite{DBLP:conf/naacl/Koncel-Kedziorski19} to encode ${G}^{t-1}$. Concretely, at step $t$ of \textit{Outlining}, graph $\mathcal{G}^{t-1}$, in the form of vertex embeddings, edge embeddings, and an adjacency matrix, is fed to the graph encoder. Here, each vertex/edge embedding is represented by the corresponding class vector, which is randomly initialized. The graph encoder establishes a pseudo graph by regarding both original vertices and edges as the nodes linked by virtual edges. For each node, a $k$-layer multi-head attention~\cite{DBLP:conf/nips/VaswaniSPUJGKP17} is performed to aggregate the information flow within its k-hop neighbors to update its own vector representation. The return consists of vertex vectors $\mathbf{V}^{t-1}=[\mathbf{v}_1,\mathbf{v}_2,...,\mathbf{v}_\mathcal{N}]$, edge vectors $\mathbf{E}^{t-1}=[\mathbf{e}_1,\mathbf{e}_2,...,\mathbf{e}_{\mathcal{N}-1}]$, and a graph vector representation $\mathbf{h}_\mathcal{G}^{t-1} \in \mathbb{R}^d$, where $\mathcal{N}$ denotes the vertex number of $\mathcal{G}^{t-1}$.

\subsection{Outlining Decoder}
We employ an LSTM as the outlining decoder, which takes the last hidden state in the encoding process as the initial state. At each step $t$ of \textit{Outlining}, it integrates the contextual information of $\mathcal{Q}$ and the structural information of $\mathcal{G}^{t-1}$ into a vector. Here, the context vector, denoted by $\mathbf{h}_\mathcal{Q}^t \in \mathbb{R}^d$, is calculated by an attention mechanism.
\begin{equation}
\label{equ:attention}
\mathbf{h}_\mathcal{Q}^t = \sum_{i=1}^\mathcal{M} {\rm softmax}({\mathbf{h}_\mathcal{G}^{t-1}}^T W_{a}\mathbf{q}_i) \mathbf{q}_i
\end{equation}
where $W_{\alpha} \in \mathbb{R}^{d \times d}$ is the trainable parameter matrix. Then, the input vector $\mathbf{h}_{in}^t$ of the AQG decoder is obtained by
\begin{equation}
\label{equ:graph_encoder}
\mathbf{h}_{in}^t = {\rm tanh}(W_{in} [\mathbf{h}_\mathcal{Q}^t; \mathbf{h}_\mathcal{G}^{t-1}])
\end{equation}
where $W_{in} \in \mathbb{R}^{d \times 2d}$ is the affine transformation.

% \subsubsection{Outlining Argument Prediction}
\noindent \textbf{Outlining Argument Prediction} 
The decoder returns the vector $\mathbf{h}_{out}^t \in \mathbb{R}^d$ and the probabilities of the $\textit{Outlining}$ arguments at step $t$ are calculated as follows.

Arguments $c_v^t$ and $s_v^t$ of \textit{AddVertex} are determined by
    \begin{equation}
    \label{equ:cv}
	c_v^t = \arg\max_{c \in \mathbf{\Psi} \cup \{\texttt{End}\}} {\rm softmax}({\rm tanh}({\mathbf{h}^t_{out}}^T W_{av}) \cdot \mathbf{x}_{c})
	\end{equation}
	\begin{equation}
    s_v^t=\left\{
    \begin{array}{lcl}
    0       &      & {t \le 4}\\
    s_v^{t-3} + \delta^t       &      & {t > 4}
    \end{array} \right.
    \end{equation}
	\begin{equation}
	\delta^t = \arg\max_{\delta \in \{0,1\}} {\rm softmax}({\rm tanh}({\mathbf{h}^t_{out}}^T W_{\delta}) \cdot \mathbf{z}_{\delta})
	\end{equation}
	where $\mathbf{x}_{c} \in \mathbb{R}^{d}$ is the embedding of vertex class $c$, $\delta^t$ denotes a signal of whether switch the current segment, and $\mathbf{z}_{\delta} \in \mathbb{R}^{d}$ is the trainable signal embedding.
    
    Argument $u$ of \textit{SelectVertex} is determined by
    \begin{equation}
	u = \arg\max_{u_i \in V^{t-1}} {\rm softmax}({\rm tanh}({\mathbf{h}^t_{out}}^T W_{sv}) \cdot \mathbf{v}_i)
	\end{equation}
	where $\mathbf{v}_{i} \in \mathbb{R}^{d}$ is the semantic vector of vertex $u_i$ obtained by the graph encoder.
	
	The argument $c_e$ of \textit{AddEdge} is determined in the same manner as (\ref{equ:cv}), while using another set of parameters.
	
To ensure that the generated structures (AQG) are syntactically correct, we mask the scores of the illegal candidate arguments at each step. For example, suppose the current operation is \textit{AddEdge}, adding an edge $e$ from vertex $u$ to $v$,  where $c_v =$ \texttt{Ent} and $c_u =$ \texttt{Var}; then, $c_e$ cannot be \texttt{Cmp}. After computing the score of each candidate $c_e$, we set the score of \texttt{Cmp} to $-\infty$. This step can be done by some simple rules.

% 	\begin{equation}
% 	c_e = \arg\max_{c \in \mathbf{\Phi}} {\rm softmax}({\rm tanh}({\mathbf{h}^t_{out}}^T W_{ae}) \cdot \mathbf{y}_{c})
% 	\end{equation}
% 	where $W_{ae} \in \mathbb{R}^{d \times d}$ is the affine transformation and $\mathbf{y}_{c} \in \mathbb{R}^{d}$ is the embedding of edge class $c$.

% \subsubsection{Copy Mechanism}
\noindent \textbf{Copy Mechanism} 
We observe that the query graph sometimes has some duplicate vertices and edges, e.g., \texttt{m.0f2y0} and \texttt{portrayed\_in\_films} in Fig. \ref{fig:query_graph}b. Although they are semantically identical, the SPARQL program repeats them for easy access. 
We would like the model to identify these vertices/edges during \textit{Outlining} to further reduce the search space; therefore, we adopt a \textit{Copy Mechanism} when performing \textit{AddVertex} and \textit{AddEdge}.
Specifically, for the vertex $v^t$ added by \textit{AddVertex} at step $t$, its semantically equivalent vertex $\tilde{v^t}$ is identified as
\begin{equation}
\label{equ:copy}
\tilde{v^t} = \arg\max_{v_i \in V^{t-1} \cup \{v_{\texttt{NONE}}\}} {\rm softmax}({\rm tanh}({\mathbf{h}^t_{out}}^T W_{cv}) \cdot \mathbf{v}_i)
\end{equation}
where $\mathbf{v}_{i}$ is obtained by the graph encoder and $v_{\texttt{NONE}}$ means that there is no such vertex. For copying edges, a similar mechanism is performed on $e^t$ of \textit{AddEdge}.

\subsection{Vertex-filling Decoder \& Edge-filling Decoder}
Consistent with the outlining decoder, both the vertex-filling decoder and the edge-filling decoder are LSTMs, starting from the last hidden state of the encoding. As described in \ref{sec:filling}, the model first decodes for filling vertices and then decodes for filling edges. 
At each decoding step $t$ of \textit{Filling}, the context vector $\mathbf{\hat{h}}_\mathcal{Q}^t \in \mathbb{R}^d$ is computed by an attention mechanism similar to that of Eq. (\ref{equ:attention}).

% At each step $t$ during \textit{Filling}, the model needs to understand the semantic role of $v^t$/$e^t$ to be filled in the entire graph, accordingly,  $\mathbf{\hat{h}}_{in}^t$, the input to the vertex filling decoder is calculated by \textit{Auxiliary Structural Encoding}.
At each step $t$ in the \textit{Filling} process, the model needs to understand the semantic role of $v^t$ to be filled in the whole graph, and accordingly, ${\hat{h}}_{in}^t$, the input to the vertex-fill decoder is calculated by the \textit{auxiliary structural encoding}.
\begin{equation}
\label{equ:auxiliary_encoding}
\mathbf{\hat{h}}_{in}^t = {\rm tanh}(\hat{W}^v_{in} [\mathbf{\hat{h}}_\mathcal{Q}^t; \mathbf{h}_{\mathcal{G}_a};
\mathbf{v}^t])
\end{equation}
% where $\mathbf{h}_{\mathcal{G}_a} \in \mathbb{R}^d$ is the vector representation of the AQG $\mathcal{G}_a$ and $\mathbf{v}^t \in \mathbb{R}^d$ is the semantic vector of $v^t$. Both $\mathbf{h}_{\mathcal{G}_a}$ and $\mathbf{v}^t$ are obtained by the graph encoder and hold auxiliary structural information. For edge filling decoder, the procedure is similar.
where $\mathbf{h}_{\mathcal{G}_a} \in \mathbb{R}^d$ is the vector representation of $\mathcal{G}_a$ and $\mathbf{v}^t \in \mathbb{R}^d$ is the semantic vector of $v^t$. Both $\mathbf{h}_{\mathcal{G}_a}$ and $\mathbf{v}^t$ are obtained by the graph encoder and hold auxiliary structural information. A similar procedure is applied to the edge-filling decoder.

\noindent \textbf{Filling Arguments Prediction} The vertex-/edge-filling decoder returns a vector $\mathbf{\hat{h}}_{out}^t \in \mathbb{R}^d$; then, the argument $i_v^t$ of \textit{FillVertex} is determined by
    \begin{equation}
    \hat{i}_v^t = \arg\max_{i \in \mathbf{I}_{c_v}} {\rm softmax}({\rm tanh}({\mathbf{\hat{h}}^t_{out}} W_{v}) \cdot \mathbf{h}_i)    
    \end{equation}
    \begin{equation}
    \label{equ:iv}
    i_v^t=\left\{
    \begin{array}{lcl}
    \hat{i}_{\tilde{v}}^t      &      & {\tilde{v^t} \neq \texttt{NONE}}\\
    \hat{i}_v^t       &      & {\tilde{v^t} = \texttt{NONE}}
    \end{array} \right.
    \end{equation}
% 	where $\mathbf{h}_i$ is the representation of instance $i$ from the instance encoder, and $\hat{i}_{\tilde{v}}^t$ is the filled instance of semantically equivalent vertex $\tilde{v^t}$. For operation \textit{FillEdge}, argument $i_e$ is determined in a similar manner.
	where $\mathbf{h}_i$ is the representation of instance $i$ from the instance encoder and $\hat{i}_{\tilde{v}}^t$ is the semantically equivalent filled instance of vertex $\tilde{v^t}$. For the operation \textit{FillEdge}, the argument $i_e$ is decided in a similar manner.
	
In our experiment, the inference process of HGNet is implemented by \textit{beam search}~\cite{DBLP:journals/corr/WisemanR16} to avoid the local optima.

\subsection{Execution Guidance Strategy}
% \noindent \textbf{Execution-Guided Strategy} 
% In general, a query graph must be illegal if its query result over the KG is empty. Motivated by this, we design an Execution-guided (EG) Strategy to avoid illegal graphs during edge \textit{Filling}, in order to reduce error propagation. Algorithm \ref{alg:eg} describes the detailed process. Briefly, at step $t$, for each candidate instance $i$, we first fill it to $\mathcal{G}^t$ (line \ref{algline:filling}) and then execute the converted SPARQL over the KG (line \ref{algline:sparql}-\ref{algline:execute}). If the result is empty, $i$ is masked by attaching $-\infty$ score for this step (line \ref{algline:mask}). Function \textsc{ToSPARQL} aims to convert $\mathcal{G}_u$ into a equivalent SPARQL. In our proposed EG, the number of execution, denoted by $\eta$, has an upper bound $\sup(\eta)= (\mathcal{N}-1) K Y_{\mathcal{I}}$, where $Y_{\mathcal{I}}$ is the total number of the candidate instances, $K$ is the beam size of each decoding step, and $\mathcal{N}$ is the vertex number of the target query graph.

In general, a query graph must be illegal if its query result on the KG $\mathcal{K}$ is empty. Inspired by this, we design an \textit{execution guidance} (EG) strategy to avoid illegal graphs during edge-\textit{Filling} and thus reduce error propagation. Algorithm \ref{alg:eg} describes the detailed procedure. Briefly, at each step $t$ of filling edges, for each candidate instance $i$, we first assume to fill it to $\mathcal{G}^t$ (line \ref{algline:filling}) and then convert $\mathcal{G}^t$ to an equivalent SPARQL $\mathcal{S}^t$ using the function \textsc{ToSPARQL}. The conversion process can be considered as the inverse process of SPARQL to the query graph (see Section \ref{sec:query_graph}). The difference is that \textsc{ToSPARQL} treats the unfilled edge slots in $\mathcal{G}^t$ as variables in $\mathcal{S}^t$ and sets the intent of $\mathcal{S}^t$ to \texttt{ASK} instead of \texttt{SELECT}. We execute $\mathcal{S}^t$ on $\mathcal{K}$ to query $\mathcal{K}$ for the existence of a graph pattern of $\mathcal{G}^t$. If the result is null, $\mathcal{G}^t$ is not legal in KG, indicating that $i$ is a wrong candidate and its predicted score will be set to $-\infty$.

Although our proposed EG requires multiple visits to KG, its number of visits $\eta$ has an upper bound $sup(\eta)= (\mathcal{N}-1) K Y_{\mathcal{I}}$, where $Y_{\mathcal{I}}$ is the total number of candidate instances, $K$ is the bundle size of each decoding step, and $\mathcal{N} -  1$ is the number of edges of the target query graph.

\begin{algorithm}[t]
	\caption{Execution Guidance Decoding for HGNet \label{alg:eg}}
	\begin{algorithmic}[1]
		\Require Question $\mathcal{Q}$. Knowledge Graph $\mathcal{K}$. AQG $\mathcal{G}_a$. Candidate instance sets  $\mathcal{I} = \{\mathcal{I}_x | x \in \mathbf{\Phi}\} $. Beam size $K$. Edges to be filled $\{e_1, e_2, ..., e_{\mathcal{N}-1}\}$. Classes of edges $\{c_1, c_2, ..., c_{\mathcal{N}-1}\}$.
		\State Initialize $t=0$, beams $\mathcal{B} = \{(\mathcal{G}_a, 0)\}$
		\While{$t<\mathcal{N}-1$}
		\State Set lived beams $\mathcal{B}_l = \emptyset$
		\For{$(\mathcal{G}^t, s_{\mathcal{G}^t}) \in \mathcal{B}$}
		\State Set time beams $\hat{\mathcal{B}} = \emptyset$
		\For{$i \in \mathcal{I}_{c_t}$}
		\State Set $\mathcal{G}_i^t = \textsc{Copy}(\mathcal{G}^t)$ and \textsc{FillEdge}($\mathcal{G}_i^t$, $e^t$, $i$)\label{algline:filling}
		\State Set $\mathcal{S}_i^t = \textsc{ToSPARQL}(\mathcal{G}_i^t)$\label{algline:sparql}
		\State $\mathcal{R}_i=\textsc{Execute}(\mathcal{S}_i^t, \mathcal{K})$\label{algline:execute}
		\If{$\mathcal{R}_i^t \neq \emptyset$}
		\State Set $s_{\mathcal{G}_i^t} = s_{\mathcal{G}^t} + \log P(\hat{i}=i|\mathcal{Q})$
		\Else
		\State Set $s_{\mathcal{G}_i^t} = s_{\mathcal{G}^t} - \infty$\label{algline:mask}
		\EndIf
		\State Set $\hat{\mathcal{B}} = \hat{\mathcal{B}} \cup \{(\mathcal{G}_i^t, s_{\mathcal{G}_i^t})\}$
		\EndFor
		\State Set $\mathcal{B}_l = \mathcal{B}_l \cup \textsc{TopK}(\hat{\mathcal{B}}, K)$
		\EndFor
		\If $\mathcal{B}_l = \emptyset$
		\State \textbf{break}
		\EndIf
		\State Set $\mathcal{B} = \textsc{TopK}(\mathcal{B}_l, K) $
		\State $t = t + 1$
		\EndWhile
		\State \Return $\textsc{TopK}(\mathcal{B}, 1)$
	\end{algorithmic} 
\end{algorithm}

\subsection{Discussion of the Search Space}
\label{sec:search_space}
% Let $Y_o$, $Y_v$, and $Y_e$ denote the average size of the argument candidate set (sub-space) at each step of \textit{Outlining}, Vertex \textit{Filling}, and Edge \textit{Filling}, respectively. Then, the total search space of generating query graph, approximated as the total number of beams, is $\hat{\mathcal{Y}} \approx (3\mathcal{N} - 1)KY_o + \mathcal{N}KY_v + (\mathcal{N} - 1)KY_e$, where beam size $K \le 5$ and vertex number $\mathcal{N}$ usually not exceeds 15.
% Generally, $Y_o, Y_v, Y_e \leq Y \approx 100$, where $Y$ is the average number of one-hop relations. Therefore, we have $\hat{\mathcal{Y}} \ll \mathcal{Y} \approx Y^L$ if $L > 3$. The beams are few enough to be embedded by the neural networks during decoding thereby challenge 2 is solved.

In the LR-based methods, the ranking model has to encode and score $\mathcal{Y}$ candidate query graphs, where $\mathcal{Y} \approx Y^L$ and $Y$ is the average number of one-hop relations in KG, and $L$ is the maximum number of hops in the core relation chain. When faced with complex NLQs, $L$ often exceeds 2, resulting in a huge number of candidate query graphs that are difficult to embed.

In contrast, our HGNet model encodes each candidate instance only once at the beginning to obtain their vector representation. In the subsequent decoding process, the total number of scoring is denoted by $\mathcal{Y}^* \approx (3\mathcal{N} - 1)KY_o + \mathcal{N}KY_v + (\mathcal{N} - 1)KY_e$. Here $K$ is the size of the beams, $3\mathcal{N} - 1$, $\mathcal{N}$, and $\mathcal{N} - 1$ denote the number of operations in the three phases \textit{Outlining}, vertex-\textit{Filling}, and edge-\textit{Filling}, respectively, and $Y_o$, $Y_v$, and $Y_e$ are the number of candidate arguments for each operation in the corresponding phases. In our experiment, we have $K = 5$, $\mathcal{N} \leq 15$, and $Y_o, Y_v, Y_e \leq Y$. Therefore, when $L > 2$, we consider that there is $\mathcal{Y}^* \ll \mathcal{Y}$. This suitable search space allows our HGNet to run on conventional devices, effectively alleviating challenge 2.

\begin{table} 
	\begin{center}
	{\caption{Recall of candidate instances.}\label{tab:instance_recall}}
	\scalebox{0.95}{
		\begin{tabular}{lccc}
			\toprule
			\textbf{,Class} & \textbf{CWQ} & \textbf{LC-QuAD} & \textbf{WebQSP}
			\\
			\cmidrule(lr){1-4}
			\texttt{Type} & 99.65 & 97.80 & - \\
			\texttt{Rel} & 95.24 & 95.32 & 93.90 \\
			\texttt{Val} & 89.63 & - & 94.82 \\
			\bottomrule
		\end{tabular}
		}
	\end{center}
\end{table}

\begin{table} 
	\begin{center}
	{\caption{Average Hit@1/F1-score on CWQ, LCQ, and WSP.}\label{tab:overall_results}}
	\scalebox{0.95}{
		\begin{tabular}{lccc}
			\toprule
			\textbf{Method} & \textbf{CWQ} & \textbf{LC-QuAD} & \textbf{WebQSP}
			\\
			\cmidrule(lr){1-4}
			STAGG~\cite{DBLP:conf/acl/YihCHG15} & -/- & -/69.0 & -/67.0 \\
			HR-BiLSTM~\cite{DBLP:conf/acl/YuYHSXZ17} & 33.3/31.2 & -/70.0 & -/68.0 \\
			GRAFT-Net~\cite{DBLP:conf/emnlp/SunDZMSC18} & 30.1/26.0 & -/- & 67.8/62.8 \\
			KBQA-GST~\cite{DBLP:conf/ijcai/LanW019} & 39.3/36.5 & -/- & 68.2/67.9 \\
			PullNet~\cite{DBLP:conf/emnlp/SunBC19} & 45.9/- & -/- & 68.1/- \\
			Slot-Matching~\cite{DBLP:conf/semweb/MaheshwariTLCF019} & -/- & -/71.0 & -/70.0 \\
			AQGNet~\cite{DBLP:conf/ijcai/ChenLHQ20} & -/- & -/74.8 & -/- \\
			DAM~\cite{DBLP:journals/kbs/ChenL20} & -/- & -/72.0 & -/70.0 \\
			QGG~\cite{DBLP:conf/acl/LanJ20} & 44.1/40.4 & -/- & -/74.0 \\
			ImprovedQGG~\cite{DBLP:conf/emnlp/QinLPA21} & -/46.2 & -/- & -/66.0 \\
			NSM+h~\cite{DBLP:conf/wsdm/HeL0ZW21} & 48.8/44.0 & -/- & 74.3/67.4 \\
			Sparse-QA~\cite{DBLP:journals/kbs/BakhshiNMR22} & -/- & 77.0/- & -/- \\
			\cmidrule(lr){1-4}
			HGNet & 65.3/64.9 & 76.0/75.1 & 71.7/71.7 \\
			\quad $+$ \textsc{Bert}-Base & \textbf{68.9}/\textbf{68.5} & \textbf{78.7}/\textbf{78.1} & \textbf{76.9}/\textbf{76.6} \\
			\bottomrule
		\end{tabular}
		}
	\end{center}
\end{table}

\subsection{Training}
% In our experiments, each training sample is a pair of an NLQ $\mathcal{Q}$ and a gold SPARQL $\mathcal{S}^+$. We train HGNet by supervised learning with teacher-forcing. During training, HGNet is optimized by maximizing the log-likelihood:
In our experiments, each training sample is a pair of NLQ $\mathcal{Q}$ and gold SPARQL $\mathcal{S}^+$. We train HGNet by supervised learning and teacher forcing. During training, HGNet is optimized by maximizing the log-likelihood:
\begin{equation}
\begin{aligned}
\mathcal{L} = &-\sum_{\mathcal{Q}} {\sum_{t=1}^{|\mathcal{A}_o|} \sum_{a \in \mathcal{A}^t_o} \log P(\hat{a}^t=a|\mathcal{Q})} \\
&- \sum_{t=1}^{|\mathcal{A}_v|}  \log P(\hat{b}^t=\mathcal{A}_v^t|\mathcal{Q}) 
- \sum_{t=1}^{|\mathcal{A}_e|}  \log P(\hat{b}^t=\mathcal{A}_e^t|\mathcal{Q})
\end{aligned}
\end{equation}
% where $\mathcal{A}_o$, $\mathcal{A}_v$, and $\mathcal{A}_e$ are the supervised signals of three generative phases, respectively. They are obtained by a \textit{Depth-First Traversal} on $\mathcal{G}^+_q$, which is the gold query graph transformed from $\mathcal{S}^+$. The traversal is started from vertex $v_{\texttt{Ans}}$, the vertex of class \texttt{Ans}, because each query graph must have a vertex denoting the answer. 
where $\mathcal{A}_o$, $\mathcal{A}_v$ and $\mathcal{A}_e$ are the supervised signals for each of the three generation stages. They are obtained from a \textit{depth-first traversal} on $\mathcal{G}^+_q$, and $\mathcal{S}^+$ is the golden query graph transformed from $\mathcal{S}^+$. The traversal starts from the answer vertex $v_{\texttt{Ans}}$ because each query graph must have a vertex representing the answer.

At the beginning of traversal, the $\texttt{Ans}$ class label is pushed into $\mathcal{A}_o$ as the gold argument of the first \textit{Outlining} operation \textit{AddVertex}, while \texttt{NONE} is pushed into $\mathcal{A}_v$ for the first \textit{Filling} operation.

Thereafter, we consider the depth-first traversal as the standard expansion process following our graph generation framework. Specifically, whenever visiting from vertex $u$ to vertex $v$ through edge $e$,  three gold \textit{Outlining} operations can be extracted, namely \textit{AddVertex}($v$), \textit{SelectVertex}($u$), and \textit{AddEdge}($e$). Correspondingly, $c_v$, $u$, and $c_{e}$ are pushed into $\mathcal{A}_o$ in sequence because they are the gold arguments. In addition, the two \textit{Filling} operations, \textit{FillVertex} and \textit{FillEdge}  can be obtained according to the \textit{AddVertex} and \textit{AddEdge}. The corresponding arguments $i_v$ and $i_{e}$ are pushed into $\mathcal{A}_v$ and $\mathcal{A}_e$, respectively.

% Once all vertices are visited, the traversal is completed. Finally, the $\texttt{End}$ label is pushed into $\mathcal{A}_o$ to represent the ending of \textit{Outlining}. Due to space limitations, the gold label of the copy mechanism and segment, and the direction of the edge are omitted here. Appendix \ref{sec:signal_building} describes the whole process in detail.
Once all vertices have been visited, the traversal is complete. Finally, the \texttt{End} tag is pushed into $\mathcal{A}_o$ to represent the end of \textit{Outlining}. Due to space limitations, the copy mechanism and gold labels of the segments, as well as the direction of the edges, are omitted here. Appendix \ref{sec:signal_building} describes the whole process in detail.

\section{Experiments}
\subsection{Experimental Setup}
Our models are trained and evaluated over three KGQA datasets: 
\textbf{WebQuestionsSP}\footnote{\url{http://aka.ms/WebQSP}} (WebQSP)~\cite{DBLP:conf/acl/YihRMCS16} contains 3,098 training and 1,639 testing NLQs that are answerable against the Freebase 2015-08 release. Most NLQs require up to 2-hop reasoning from the KG with constraints of entities and time.
\textbf{LC-QuAD}\footnote{\url{https://figshare.com/projects/LC-QuAD/21812}}~\cite{DBLP:conf/semweb/TrivediMDL17} is a gold standard complex question answering benchmark over the DBpedia 2016-04 release, having 3,500 training, 500 validation, and 1,000 testing pairs of NLQ and SPARQL queries. It contains three types of NLQs: selection, Boolean, and count.
\textbf{ComplexWebQuestions}\footnote{\url{https://www.tau-nlp.org/compwebq}} (CWQ)~\cite{DBLP:conf/naacl/TalmorB18} is currently one of the hardest KGQA benchmarks, and each NLQ corresponds to an executable SPARQL query. It has 27,623 training, 3,518 validation, and 3,531 testing pairs of NLQ and SPARQL queries. These NLQs require up to 4-hop reasoning with some complicated constraints of comparison, ordinary and nested queries.
    
% For all datasets, we adopt Precision, Recall, F1-score, and Hit@1 as the evaluation metrics, which are consistent with the previous work~\cite{DBLP:conf/emnlp/Hu0Z18,DBLP:conf/emnlp/SunBC19,DBLP:conf/ijcai/ChenLHQ20,DBLP:conf/wsdm/HeL0ZW21}. We also evaluate our method in terms of AQG Accuracy ($\mathcal{G}_a$ Acc.) and Query Graph Accuracy ($\mathcal{G}_q$ Acc.), which directly reflects the performance on semantic understanding.
For all datasets, we use Precision, Recall, F1-score, and Hit@1 as evaluation metrics, which is consistent with previous work~\cite{DBLP:conf/emnlp/Hu0Z18,DBLP:conf/emnlp/SunBC19,DBLP:conf/ijcai/ChenLHQ20,DBLP:conf/wsdm/HeL0ZW21}. We also evaluate our method in terms of AQG accuracy ($\mathcal{G}_a$ Acc.) and query graph accuracy ($\mathcal{G}_q$ Acc.), which directly reflect the performance in terms of semantic understanding.

\noindent \textbf{Implementation Details} In our experiments, all word embeddings are initialized with 300-d pre-trained word embeddings using GloVe~\cite{DBLP:conf/emnlp/PenningtonSM14}. The hyperparameters of HGNet are set as follows: (1) The dimension of all the semantic vectors, i.e., $d_h$, is set to 256; (2) The layer number of all the BiLSTMs is set to 1, and the layer number of the graph transformer is set to 3; (3) The learning rate is set to $2 \times 10^{-4}$; (4) The beam size is set to 5. (5) The batch size is set to 16. (6) For classes \texttt{Rel} and \texttt{Type},  the numbers of candidate instances are set to 50 and 3, respectively. (7) The beam sizes of all decoding are set to 5.  All our codes are publicly available\footnote{\url{https://github.com/Bahuia/HGNet}}.

\noindent \textbf{Performance of Candidate Instance Collecting} Table \ref{tab:instance_recall} gives the recall for the candidate sets of \texttt{Type}, \texttt{Rel}, and \texttt{Val}. Some classes of \texttt{Ent}, \texttt{Ord}, \texttt{Cmp}, and \texttt{Agg} are not shown because they have a recall of 100\%. In addition, WebQSP has no \texttt{Type} instances and LC-QuAD has no \texttt{Val} instances.

\noindent \textbf{Methods for Comparison}
We first compared our method with the following LR-based methods, including STAGG~\cite{DBLP:conf/acl/YihCHG15}, HR-BiLSTM~\cite{DBLP:conf/acl/YuYHSXZ17}, and DAM~\cite{DBLP:journals/kbs/ChenL20}, Slot-Matching~\cite{DBLP:conf/semweb/MaheshwariTLCF019}, and our previous work, AQGNet~\cite{DBLP:conf/ijcai/ChenLHQ20}. Here AQGNet utilizes the AQG to constrain the ranking of the query graph, which is the basis of this paper. However, it cannot handle CWQ because it does not consider the complex SPARQL and follows the query ranking. Because of the lack of supervised methods for CWQ, we must compare with weakly supervised methods: KBQA-GST~\cite{DBLP:conf/ijcai/LanW019}, Sparse-QA~\cite{DBLP:journals/kbs/BakhshiNMR22}, QGG~\cite{DBLP:conf/acl/LanJ20},
ImprovedQGG~\cite{DBLP:conf/emnlp/QinLPA21},GRAFT-Net~\cite{DBLP:conf/emnlp/SunDZMSC18}, PullNet~\cite{DBLP:conf/emnlp/SunBC19}, and NSM+h~\cite{DBLP:conf/wsdm/HeL0ZW21}. Although they dropped the SPARQL annotation, they still extracted paths or subgraphs from the KG as supervised signals for inference.

\noindent \textbf{Equipped with PLM}
To explore whether a pre-trained language model (PLM) would be helpful for HGNet, we modified the base HGNet to a more advanced version: replacing the NLQ encoder and the instance encoder with \textsc{Bert}-base~\cite{DBLP:conf/naacl/DevlinCLT19} and fine-tuning them, keeping the rest of the model unchanged.

% \begin{figure*}
%   \centering
%     \begin{subfigure}{0.3\textwidth}
%       \centering   
%       \includegraphics[width=1\linewidth]{figure/type_cwq.pdf}
%         \caption{Results on CWQ}
%         \label{fig:type_cwq}
%     \end{subfigure}   %      \hfill  %
%     \begin{subfigure}{0.3\textwidth}
%       \centering   
%       \includegraphics[width=\linewidth]{figure/type_lcq.pdf}
%         \caption{Results on LC-QuAD}
%         \label{fig:type_lcq}
%     \end{subfigure}
%     \begin{subfigure}{0.3\textwidth}
%       \centering   
%       \includegraphics[width=\linewidth]{figure/type_wsp.pdf}
%         \caption{Results on WebQSP}
%         \label{fig:type_wsp}
%     \end{subfigure}
% \caption{
% \label{fig:type}
% F1-score on different types of SPARQL syntax.
% }
% \end{figure*}

\begin{table*} 
	\begin{center}
	{\caption{Experimental results for comparison with baselines.}\label{tab:baselines}}
	\scalebox{0.95}{
		\begin{tabular}{lcccccccccccc}
			\toprule
			\multicolumn{1}{l}{\multirow{2}[1]{*}{\textbf{Baselines}}}
			&\multicolumn{4}{c}{\textbf{CWQ}}&\multicolumn{4}{c}{\textbf{LC-QuAD}}&\multicolumn{4}{c}{\textbf{WebQSP}}\\
		    \cmidrule(lr){2-5} \cmidrule(lr){6-9} \cmidrule(lr){10-13}
			
			& $\mathcal{G}_q$ Acc. &Prec. &Rec. &F1 
			& $\mathcal{G}_q$ Acc. &Prec. &Rec. &F1 
			& $\mathcal{G}_q$ Acc. &Prec. &Rec. &F1 \\

			\cmidrule(lr){1-1} \cmidrule(lr){2-5} \cmidrule(lr){6-9} \cmidrule(lr){10-13}
			\textsc{Bart}-Base &- &27.97 &28.30 &27.62 &- &48.01 &49.19 &47.62 &- &53.20 &56.17 &53.49 \\
			LR &- &- &- &- &54.75 &65.89 &75.30 &69.53 &62.58 &70.82 &\textbf{80.50} &71.52 \\
			\quad $+$ \textsc{Bert}-Base &- &- &- &- &58.03 &69.71 &77.14 &72.60 &65.38 &74.09 &\textbf{81.83} &75.42 \\
			StrLR &- &- &- &- &60.80 &75.54 &74.95 &74.81 &61.95 &70.30 &74.32 &70.89 \\
			\quad $+$ \textsc{Bert}-Base &- &- &- &- &63.25 &78.27 &77.36 &77.87 &64.74 &74.31 &78.59 &75.06 \\
			NHGG &47.89 &59.09 &63.15 &59.12 &27.10 &46.93 &48.36 &46.12 &54.59 &60.74 &64.22 &60.68 \\ 
			\cmidrule(lr){1-1} \cmidrule(lr){2-5} \cmidrule(lr){6-9} \cmidrule(lr){10-13}
			HGNet &54.59 &65.27 &68.44 &64.95 &60.90 &75.82 &75.22 &75.10 &66.19 &71.58 &75.10 &71.71 \\ 
			\quad $+$ \textsc{Bert}-Base &\textbf{57.80} &\textbf{68.89} &\textbf{73.30} &\textbf{68.88} &\textbf{63.50} &\textbf{78.92} &\textbf{78.14} &\textbf{78.13} &\textbf{70.74} &\textbf{76.66} &79.28 &\textbf{76.62} \\ 
			\bottomrule
		\end{tabular}
		}
	\end{center}
\end{table*}

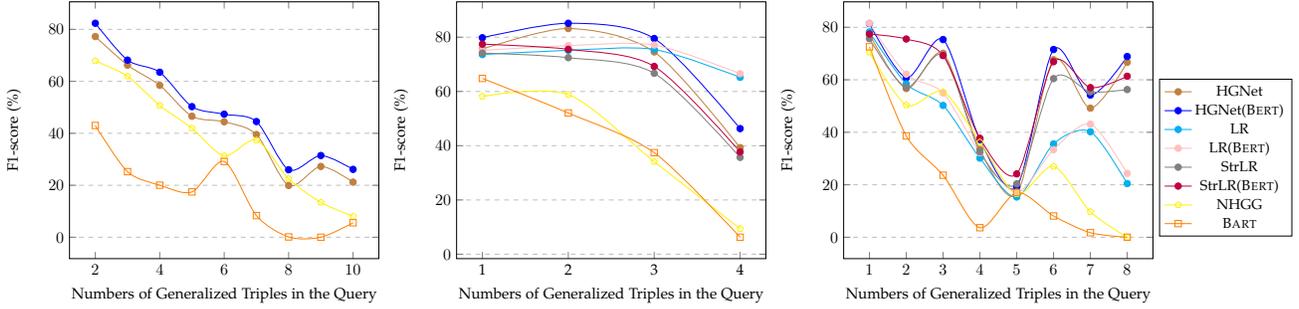
\begin{figure*}
% \vspace{-3mm}
\centering
    \begin{tikzpicture}[scale=0.6]
        \begin{axis}[
        xlabel={Numbers of Generalized Triples in the Query}, ylabel={F1-score (\%)}, 
        % tick align=inside, 
        %legend pos=south west, 
        % legend entries = none,
        % legend style={font=\small}, 
        ymajorgrids=true, 
        grid style=dashed]

            \addplot[smooth,mark=*,brown] plot coordinates { 
                (2, 77.24) (3, 66.23) (4, 58.46) (5, 46.56) (6, 44.43) (7, 39.52) (8, 19.92) (9, 27.25) (10, 21.21)
            };
            % \addlegendentry{HGNet}
        
            \addplot[smooth,mark=*,blue] plot coordinates {
                (2, 82.35) (3, 68.17) (4, 63.51) (5, 50.27) (6, 47.35) (7, 44.52) (8, 25.98) (9, 31.45) (10, 26.12)
            };
            % \addlegendentry{HGNet(\textsc{Bert})}
            
            \addplot[smooth,mark=pentagon,yellow] plot coordinates {
                (2, 67.83) (3, 61.97) (4, 50.72) (5, 41.97) (6, 31.31) (7, 37.39) (8, 22.54) (9, 13.44) (10, 8.05)
            };
            % \addlegendentry{NHGG}
            
            \addplot[smooth,mark=square,orange] plot coordinates {
                (2, 43.01) (3, 25.23) (4, 20.04) (5, 17.45) (6, 29.17) (7, 8.33) (8, 0.12) (9, 0 ) (10, 5.56)
            };
            % \addlegendentry{\textsc{Bart}}
            
        \end{axis}
    \end{tikzpicture}
    % \qquad
    \begin{tikzpicture}[scale=0.6]
        \begin{axis}[xlabel={Numbers of Generalized Triples in the Query},
        ylabel={F1-score (\%)}, 
        xtick=data,
        xticklabels={1, 2, 3, 4},
        tick align=inside, 
        legend pos=south west, 
        legend style={font=\small}, 
        ymajorgrids=true, 
        grid style=dashed]

            \addplot[smooth,mark=*,brown] plot coordinates { 
                (1, 75.60) (2, 83.19) (3, 74.57) (4, 39.29)
            };
            % \addlegendentry{HGNet}
        
            \addplot[smooth,mark=*,blue] plot coordinates {
                (1, 79.80) (2, 85.14) (3, 79.52) (4, 46.30)
            };
            % \addlegendentry{HGNet(\textsc{Bert})}
            
            \addplot[smooth,mark=*,cyan] plot coordinates {
                (1, 73.59) (2, 75.16) (3, 75.48) (4, 65.21)
            };
            % \addlegendentry{LR}
            
            \addplot[smooth,mark=*,pink] plot coordinates {
                (1, 75.12) (2, 76.90) (3, 77.11) (4, 66.54)
            };
            % \addlegendentry{LR(\textsc{Bert})}
            
            \addplot[smooth,mark=*,gray] plot coordinates {
                (1, 74.15) (2, 72.45) (3, 66.70) (4, 35.69)
            };
            % \addlegendentry{StrLR}
            
            \addplot[smooth,mark=*,purple] plot coordinates {
                (1, 77.43) (2, 75.54) (3, 69.23) (4, 37.72)
            };
            % \addlegendentry{StrLR(\textsc{Bert})}
            
            \addplot[smooth,mark=pentagon,yellow] plot coordinates {
                (1, 58.18) (2, 58.85) (3, 34.13) (4, 9.38)
            };
            % \addlegendentry{NHGG}
            
            \addplot[smooth,mark=square,orange] plot coordinates {
                (1, 64.81) (2, 52.06) (3, 37.44) (4, 6.25)
            };
            % \addlegendentry{\textsc{Bart}}
            
        \end{axis}
    \end{tikzpicture}
    \begin{tikzpicture}[scale=0.6]
        \begin{axis}[xlabel={Numbers of Generalized Triples in the Query},
        ylabel={F1-score (\%)}, 
        xtick=data,
        tick align=inside, 
        legend style={font=\small, at={(1.45,0.7)}},
        ymajorgrids=true, 
        grid style=dashed]

            \addplot[smooth,mark=*,brown] plot coordinates { 
                (1, 77.49) (2, 56.67) (3, 70.10) (4, 33.46) (5, 16.47) (6, 67.64) (7, 49.19) (8, 66.67)
            };
            \addlegendentry{HGNet}
        
            \addplot[smooth,mark=*,blue] plot coordinates {
                (1, 81.43) (2, 60.78) (3, 75.33) (4, 37.21) (5, 19.23) (6, 71.54) (7, 54.18) (8, 68.88)
            };
            \addlegendentry{HGNet(\textsc{Bert})}
            
            \addplot[smooth,mark=*,cyan] plot coordinates {
                (1, 78.24) (2, 58.45) (3, 50.23) (4, 30.15) (5, 15.34) (6, 35.56) (7, 40.20) (8, 20.45)
            };
            \addlegendentry{LR}
            
            \addplot[smooth,mark=*,pink] plot coordinates {
                (1, 81.54) (2, 62.18) (3, 54.99) (4, 32.21) (5, 17.12) (6, 33.36) (7, 43.10) (8, 24.32)
            };
            \addlegendentry{LR(\textsc{Bert})}
            
            \addplot[smooth,mark=*,gray] plot coordinates {
                (1, 75.67) (2, 57.21) (3, 69.34) (4, 32.57) (5, 20.34) (6, 60.45) (7, 55.42) (8, 56.28)
            };
            \addlegendentry{StrLR}
            
            \addplot[smooth,mark=*,purple] plot coordinates {
                (1, 77.43) (2, 75.54) (3, 69.23) (4, 37.72) (5, 24.14) (6, 66.88) (7, 57.03) (8, 61.36)
            };
            \addlegendentry{StrLR(\textsc{Bert})}
            
            \addplot[smooth,mark=pentagon,yellow] plot coordinates {
                (1, 70.48) (2, 50.36) (3, 55.41) (4, 36.31) (5, 16.47) (6, 27.01) (7, 9.71) (8, 0.0)
            };
            \addlegendentry{NHGG}
            
            \addplot[smooth,mark=square,orange] plot coordinates {
               (1, 72.57) (2, 38.62) (3, 23.64) (4, 3.63) (5, 17.04) (6, 8.10 ) (7, 1.73) (8, 0.0)
            };
            \addlegendentry{\textsc{Bart}}
            
        \end{axis}
    \end{tikzpicture}

\caption{F1-score on different complexity levels of NLQs in CWQ, LC-QuAD, and WebQSP.}
\label{fig:level}
\end{figure*}

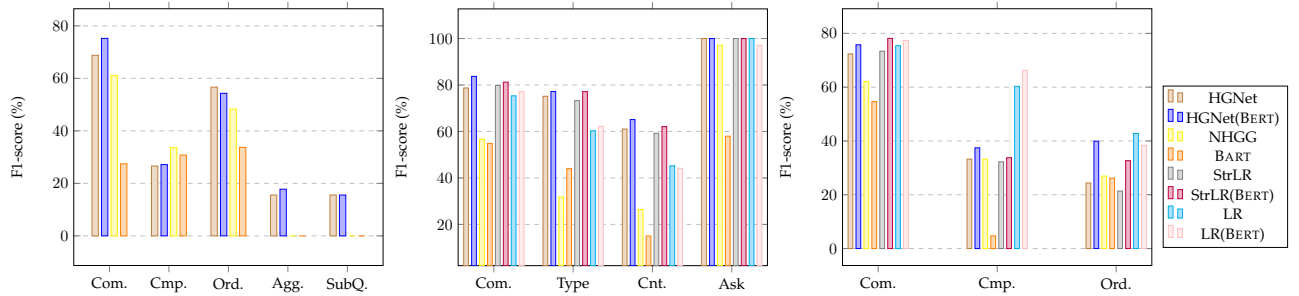
\begin{figure*}
% \vspace{-3mm}
\centering
    \begin{tikzpicture}[scale=0.6]
        \begin{axis}[
        ybar,
        bar width=4pt,
        enlargelimits=0.15,
        legend style={at={(0.5,-0.15)},
          anchor=north,legend columns=-1},
        ylabel=F1-score (\%),
        symbolic x coords={Com.,Cmp.,Ord.,Agg.,SubQ.},
    	xtick=data,
        ymajorgrids=true, 
        grid style=dashed]
    	
        \addplot[draw=brown, fill=brown!30!white] 
    	coordinates {(Com.,68.79) (Cmp.,26.59) (Ord.,56.66) (Agg.,15.56) (SubQ.,15.56)};
    	
    	\addplot[draw=blue, fill=blue!30!white]  
    	coordinates {(Com.,75.24) (Cmp.,27.15) (Ord.,54.31) (Agg.,17.78) (SubQ.,15.56)};
    
        \addplot[draw=yellow, fill=yellow!30!white]  
    	coordinates {(Com.,61.06) (Cmp.,33.60) (Ord.,48.29) (Agg.,0) (SubQ.,0)};
    
        \addplot[draw=orange, fill=orange!30!white]   
    	coordinates {(Com.,27.45) (Cmp.,30.78) (Ord.,33.72) (Agg.,0) (SubQ.,0)};
    
        % \legend{$\sigma=0.01$,$\sigma=0.05$,$\sigma=0.1$,$\sigma=0.2$}
    \end{axis}
    \end{tikzpicture}
    % \qquad
    \begin{tikzpicture}[scale=0.6]
        \begin{axis}[
        ybar,
        bar width=3pt,
        enlargelimits=0.15,
        legend style={at={(0.5,-0.15)},
          anchor=north,legend columns=-1},
        ylabel=F1-score (\%),
        symbolic x coords={Com.,Type,Cnt.,Ask},
    	xtick=data,
        ymajorgrids=true, 
        grid style=dashed]
    	
        \addplot[draw=brown, fill=brown!30!white]
    	coordinates {(Com.,78.71) (Type,75.14) (Cnt.,61.03) (Ask,100)};
    	
    	\addplot[draw=blue, fill=blue!30!white] 
    	coordinates {(Com.,83.71) (Type,77.21) (Cnt.,65.12) (Ask,100)};
    
        \addplot[draw=yellow, fill=yellow!30!white] 
    	coordinates {(Com.,56.65) (Type,31.78) (Cnt.,26.43) (Ask,97.1)};
    
        \addplot[draw=orange, fill=orange!30!white]  
    	coordinates {(Com.,54.81) (Type,44.02) (Cnt.,15) (Ask,57.97)};
    	
    	\addplot[draw=gray, fill=gray!30!white] 
    	coordinates {(Com.,79.81) (Type,73.24) (Cnt.,59.24) (Ask,100)};
    	
    	\addplot[draw=purple, fill=purple!30!white]  
    	coordinates {(Com.,81.23) (Type,77.15) (Cnt.,62.08) (Ask,100)};
    	
    	\addplot[draw=cyan, fill=cyan!30!white]  
    	coordinates {(Com.,75.34) (Type,60.3) (Cnt.,45.24) (Ask,100)};
    	
    	\addplot[draw=pink, fill=pink!30!white]   
    	coordinates {(Com.,77.19) (Type,62.23) (Cnt.,44.13) (Ask,97.1)};
    
        % \legend{$\sigma=0.01$,$\sigma=0.05$,$\sigma=0.1$,$\sigma=0.2$}
    \end{axis}
    \end{tikzpicture}
    \begin{tikzpicture}[scale=0.6]
        \begin{axis}[
        ybar,
        bar width=3pt,
        enlargelimits=0.15,
        legend style={at={(1.44,0.7)}},
        ylabel=F1-score (\%),
        symbolic x coords={Com.,Cmp.,Ord.},
    	xtick=data,
        ymajorgrids=true, 
        grid style=dashed]
    	
        \addplot[draw=brown, fill=brown!30!white]
    	coordinates {(Com.,72.39) (Cmp.,33.24) (Ord.,24.39)};
    	
    	\addplot[draw=blue, fill=blue!30!white]
    	coordinates {(Com.,75.76) (Cmp.,37.44) (Ord.,39.92)};
    
        \addplot[draw=yellow, fill=yellow!30!white]
    	coordinates {(Com.,62.04) (Cmp.,33.24) (Ord.,26.94)};
    
        \addplot[draw=orange, fill=orange!30!white] 
    	coordinates {(Com.,54.64) (Cmp.,4.71) (Ord.,26.19)};
    	
    	\addplot[draw=gray, fill=gray!30!white]  
    	coordinates {(Com.,73.39) (Cmp.,32.24) (Ord.,21.39)};
    	
    	\addplot[draw=purple, fill=purple!30!white]  
    	coordinates {(Com.,78.15) (Cmp.,33.78) (Ord.,32.67)};
    	
    	\addplot[draw=cyan, fill=cyan!30!white] 
    	coordinates {(Com.,75.41) (Cmp.,60.32) (Ord.,42.78)};
    	
    	\addplot[draw=pink, fill=pink!30!white]  
    	coordinates {(Com.,77.39) (Cmp.,66.24) (Ord.,38.39)};
    
        \legend{HGNet,HGNet(\textsc{Bert}),NHGG,\textsc{Bart},StrLR,StrLR(\textsc{Bert}),LR,LR(\textsc{Bert})}
    \end{axis}
    \end{tikzpicture}

\caption{F1-score on different types of SPARQL syntax in CWQ, LC-QuAD, and WebQSP.}
\label{fig:type}
\end{figure*}

\subsection{Overall Results}

% The experimental results are reported in Table \ref{tab:overall_results}.  Our HGNet achieves the state-of-the-art on the hardest CWQ and significantly improves the performance by 16.5\% and 20.9\% in terms of hit@1 and F1-score, respectively. On LC-QuAD and WebQSP, HGNet ranks second and third, respectively, while still outperforming all the ranking-based methods~\cite{DBLP:conf/semweb/MaheshwariTLCF019,DBLP:conf/ijcai/ChenLHQ20}. Even more exciting, using \textsc{Bert}-base as the encoder, the performance of HGNet is further improved, achieving state-of-the-art results on all three datasets. 
The experimental results are shown in Table \ref{tab:overall_results}. Our HGNet achieves the most advanced level on the most difficult CWQ and substantially improves the performance in terms of hit@1 and F1 scores by 16.5\% and 20.9\%, respectively. On LC-QuAD and WebQSP, HGNet ranked second and third, respectively, while still outperforming all ranking-based methods~\cite{DBLP:conf/semweb/MaheshwariTLCF019,DBLP:conf/ijcai/ChenLHQ20}. More excitingly, using \textsc{Bert}-base as the encoder, HGNet's performance is further improved and achieves state-of-the-art results on all three datasets.

Weakly supervised methods~\cite{DBLP:conf/ijcai/LanW019,DBLP:conf/acl/LanJ20,DBLP:conf/emnlp/SunBC19,DBLP:conf/emnlp/SunDZMSC18,DBLP:conf/wsdm/HeL0ZW21} perform poorly on CWQ because they cannot learn fine-grained question semantics without detailed SPARQL supervision. Our query graph provides a paradigm for training models using complex SPARQL. The naive HGNet does not perform as well as Sparse-QA on LC-QuAD because the latter introduces external knowledge to build a relation pattern dictionary to help with relation prediction.
It also does not achieve SOTA results on WebQSP because 95\% of SPARQL queries in WebQSP do not have more than two-hop relational chains, and the complex SPARQL syntax prevents our method from playing to its strengths. However, \textsc{Bert} compensates for these deficiencies by enhancing semantic understanding, which eventually allows our method to win over all rivals.

% \begin{figure*}
%   \centering
%     \begin{subfigure}{0.3\textwidth}
%       \centering   
%       \includegraphics[width=1\linewidth]{figure/few_shot_cwq.pdf}
%         \caption{Results on CWQ}
%         \label{fig:few_shot_cwq}
%     \end{subfigure}   %      \hfill  %
%     \begin{subfigure}{0.3\textwidth}
%       \centering   
%       \includegraphics[width=\linewidth]{figure/few_shot_lcq.pdf}
%         \caption{Results on LC-QuAD}
%         \label{fig:few_shot_lcq}
%     \end{subfigure}
%     \begin{subfigure}{0.3\textwidth}
%       \centering   
%       \includegraphics[width=\linewidth]{figure/few_shot_wsp.pdf}
%         \caption{Results on WebQSP}
%         \label{fig:few_shot_wsp}
%     \end{subfigure}
% \caption{
% \label{fig:few_shot}
% Accuracy of query graph construction with different proportions of training data.
% }
% \end{figure*}

\begin{table*} 
	\begin{center}
	{\caption{Experimental results of ablation test.\label{tab:ablation_test}}}
	\scalebox{0.95}{
		\begin{tabular}{lccccccccc}
			\toprule
			\multicolumn{1}{l}{\multirow{2}[1]{*}{\textbf{Settings}}}
			&\multicolumn{3}{c}{\textbf{CWQ}}&\multicolumn{3}{c}{\textbf{LC-QuAD}}&\multicolumn{3}{c}{\textbf{WebQSP}}\\
			\cmidrule(lr){2-4} \cmidrule(lr){5-7} \cmidrule(lr){8-10}
			
			& $\mathcal{G}_a$ Acc. &$\mathcal{G}_q$ Acc. &F1 
			& $\mathcal{G}_a$ Acc. &$\mathcal{G}_q$ Acc. &F1 
			& $\mathcal{G}_a$ Acc. &$\mathcal{G}_q$ Acc. &F1 \\
		    \cmidrule(lr){1-1} \cmidrule(lr){2-4} \cmidrule(lr){5-7} \cmidrule(lr){8-10}
			HGNet &67.13 &\textbf{54.59} &64.95 &78.00 &\textbf{60.90} &\textbf{75.10} &\textbf{82.41} &\textbf{66.19} &\textbf{71.71} \\
			\cmidrule(lr){1-1} \cmidrule(lr){2-4} \cmidrule(lr){5-7} \cmidrule(lr){8-10}
			$-$ Graph Encoder &61.03 &47.75 &61.77 &70.70 &56.70 &70.78 &73.42 &58.64 &66.82 \\
			%w/o segment embedding &66.06 &51.56 &\textbf{65.62} &- &- &- &- &- &- \\
			$-$ Auxiliary Encoding &66.90 &49.22 &62.81 &\textbf{78.10} &60.00 &73.63 &78.91 &60.45 &68.41 \\
% 			\quad \quad graph auxiliary encoding &66.41 &\textbf{52.31} &65.34 &77.10 &59.7 &72.8 &78.85 &61.82 &70.48 \\
% 			\quad \quad vertex auxiliary encoding &66.06 &50.92 &64.96 &75.90 &57.70 &71.06 &78.04 &61.88 &70.25 \\
% 			\quad \quad edge auxiliary encoding &67.18 &52.22 &65.60 &76.40 &\textbf{61.10} &73.25 &78.04 &61.07 &70.12 \\
			$-$ Copy &66.35 &50.32 &\textbf{65.97} &76.70 &52.90 &71.00 &79.23 &63.76 &71.44 \\
			$-$ EG &\textbf{69.58} &44.11 &50.96 &76.90 &32.30 &35.91 &80.16 &57.33 &61.88 \\
			Repl. LSTM with Transformer &65.82 &52.19 &62.75 &77.10 &60.30 &74.90 &81.89 &65.72 &70.98 \\
% 			\quad \quad vertex copy &65.51 &49.68 &64.47 &\textbf{78.40} &53.90 &71.87 &76.54 &61.44 &70.34 \\
% 			\quad \quad edge copy &66.75 &51.70 &\textbf{66.34} &75.80 &58.30 &71.26 &79.10 &61.32 &69.47 \\
			\bottomrule
		\end{tabular}
		}
	\end{center}
\end{table*}

\begin{figure*}
% \vspace{-3mm}
\centering
    \begin{tikzpicture}[scale=0.6]
        \begin{axis}[
        xlabel={Proption of training data (\%)}, 
        ylabel={$\mathcal{G}_q$ Acc. (\%)}, 
        xtick=data,
        tick align=inside, 
        %legend pos=south west, 
        % legend entries = none,
        % legend style={font=\small}, 
        ymajorgrids=true, 
        grid style=dashed]

            \addplot[smooth,mark=*,red] plot coordinates { 
                (1, 17.04) (5, 27.44) (10, 38.39) (30, 47.00)
            };
            % \addlegendentry{HGNet}
        
            \addplot[smooth,mark=*,blue] plot coordinates {
                (1, 16.14) (5, 27.47) (10, 35.79) (30, 45.17)
            };
            % \addlegendentry{HGNet(\textsc{Bert})}
            
            \addplot[smooth,mark=pentagon,yellow] plot coordinates {
                (1, 11.67) (5, 22.96) (10, 33.16 ) (30, 42.93)
            };
            % \addlegendentry{NHGG}
            
            \addplot[smooth,mark=square,orange] plot coordinates {
                (1, 10.05) (5, 23.34) (10, 31.60) (30, 40.70)
            };
            % \addlegendentry{\textsc{Bart}}
            
        \end{axis}
    \end{tikzpicture}
    % \qquad
    \begin{tikzpicture}[scale=0.6]
        \begin{axis}[xlabel={Proption of training data (\%)}, ylabel={$\mathcal{G}_q$ Acc. (\%)}, 
        xtick=data,
        tick align=inside, 
        legend pos=south west, 
        legend style={font=\small}, 
        ymajorgrids=true, 
        grid style=dashed]

            \addplot[smooth,mark=*,red] plot coordinates { 
                (5, 30.1) (10, 46.5) (30, 53.7) (50, 54.8)
            };
            % \addlegendentry{HGNet}
        
            \addplot[smooth,mark=*,blue] plot coordinates {
                 (5, 35.4) (10, 39.1) (30, 45.5) (50, 48.9)
            };
            % \addlegendentry{HGNet(\textsc{Bert})}
            
            \addplot[smooth,mark=pentagon,yellow] plot coordinates {
                (5, 44.4) (10, 46.4) (30, 55.4) (50, 58)
            };
            % \addlegendentry{NHGG}
            
            \addplot[smooth,mark=square,orange] plot coordinates {
                 (5, 36.6) (10, 43.8) (30, 50.1) (50, 51.6)
            };
            % \addlegendentry{\textsc{Bart}}
            
        \end{axis}
    \end{tikzpicture}
    \begin{tikzpicture}[scale=0.6]
        \begin{axis}[xlabel={Proption of training data (\%)}, ylabel={$\mathcal{G}_q$ Acc. (\%)}, 
        xtick=data,
        tick align=inside, 
        legend style={font=\small, at={(1.1,0.4)}},
        ymajorgrids=true, 
        grid style=dashed]

            \addplot[smooth,mark=*,red] plot coordinates { 
                (5, 30.44) (10, 34.62) (30, 52.59) (50, 52.59)
            };
            \addlegendentry{HGNet}
        
            \addplot[smooth,mark=*,blue] plot coordinates {
                 (5, 27.95) (10, 33.44) (30, 49.03) (50, 55.4 )
            };
            \addlegendentry{$-$ Copy}
            
            \addplot[smooth,mark=pentagon,yellow] plot coordinates {
                (5, 27.51) (10, 33.62) (30, 50.03) (50, 52.78)
            };
            \addlegendentry{$-$ AuxiliaryEncoding}
            
            \addplot[smooth,mark=square,orange] plot coordinates {
                 (5, 26.95) (10, 28.51) (30, 46.72) (50, 53.65)
            };
            \addlegendentry{$-$ GraphEncoder}
            
        \end{axis}
    \end{tikzpicture}

\caption{Accuracy of query graph construction with different proportions of training data in CWQ, LC-QuAD, and WebQSP.}
\label{fig:few_shot}
\end{figure*}
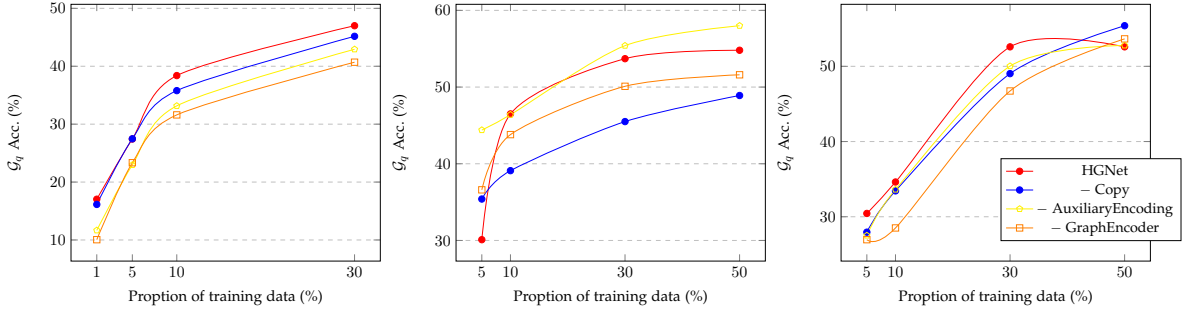

\subsection{Detailed Results and Analysis}

\subsubsection{Comparison with Baselines of Supervised Learning}
We implemented several supervised baselines for a fairer comparison.

    \noindent \textbf{\textsc{Bart}}~\cite{DBLP:conf/acl/LewisLGGMLSZ20}, which is a strong pre-trained sequence-to-sequence model, directly treats the problem as a conventional machine translation task from NLQ to SPARQL.
    
    \noindent \textbf{Learn-to-Rank} (LR) first generates candidate query graphs by STAGG~\cite{DBLP:conf/acl/YihCHG15} and then ranks the candidate graphs with a strong ranking model CompQA~\cite{DBLP:conf/emnlp/LuoLLZ18}.
    
     \noindent \textbf{LR with structural constraints} (StrLR) generates AQG $\mathcal{G}_a$ by \textit{Outlining} and subsequently generates candidate graphs by enumerating combinations of instances to populate $\mathcal{G}_a$. Thereafter, the candidate graphs are ranked using CompQA.
     
     \noindent \textbf{Non-hierarchical Graph Generation} (NHGG) integrates \textit{Outlining} and \textit{Filling} into a single procedure. For \textit{AddVertex} and \textit{AddEdge}, the model predicts instances directly instead of classes. In this way, the query graph can be completed by a single decoding procedure without the \textit{Filling} operation.

To make the baseline more robust, we also try to enhance the ranking models of LR and StrLR with \textsc{Bert}-base.

Table \ref{tab:baselines} displays the experimental results. LR and StrLR have no results on CWQ because they cannot cope with challenge 2), even though the search space of the latter has been reduced by using AQG. Moreover, \textsc{Bart} has no results for $\mathcal{G}_q$ Acc as it merely generates SPARQL. Note that our HGNet has better $\mathcal{G}_q$ than LR on WebQSP but performs worse in terms of precision, recall and F1-score. This is due to the problem of pseudo query graphs, i.e., LR obtains some incorrect query graphs that have the wrong semantics but coincidentally retrieve the correct answer.
This phenomenon is definitely not expected from a good system, so we focus on the performance on $\mathcal{G}_q Acc$. The results show that our HGNet outperforms all baselines in terms of $\mathcal{G}_q$ Acc on all datasets. Even when compared with StrLR, which also utilizes structural constraints, HGNet is still competitive ($+$0.1\% on LC-QuAD, $+$0.05\% on WebQSP). This demonstrates that our \textit{Filling} process is comparable to \textit{Ranking} in terms of performance while significantly reducing the space cost. In terms of $\mathcal{G}_q$, our substantial improvements to NHGG (3.7\%, 33.8\%, 8.0\%) reveal the need for a hierarchical procedure. Here, the extremely poor performance of NHGG on LC-QuAD is mainly attributed to the considerable amount of \texttt{Type} in this benchmark. Without \textit{Outlining}, it is difficult for the model to select the only correct \texttt{Type} from the full pool of candidate instances containing all categories.BART does not work well on all datasets, especially CWQ, because it ignores the structural features of SPARQL and flattens it into a sequence of tokens. In this way, any wrong token leads to failure of the whole SPARQL. Finally, we find that {Bert} brings 3-5\% improvement, whether for HGNet, LR or StrLR, which again proves the effectiveness of generic PLM in semantic understanding.

\begin{table*} 
	\begin{center}
	{\caption{Major causes for failures and descriptions on the test set of CWQ.}\label{tab:errors}}
	    \scalebox{0.95}{
		\begin{tabular}{llc}
			\toprule
			\textbf{Cause of Error} & \textbf{Description} & \textbf{Proportion}
			\\
			\hline
			\\[-6pt]
			Comparison Constraint
			& \makecell[l]{The \texttt{FILTER}-clauses that denote a comparison relationship in $\mathcal{S}_p$ are redundant or missing.} 
			& 18\% 
			\\
			Ordinal Constraint 
			& The \texttt{ORDER BY}-clause in $\mathcal{S}_p$ is missing or its variable object is incorrect.  
			& 2\%  
			\\
			Relation Triple 
			& The relation triples in $\mathcal{S}_p$ are redundant or missing, or their topology is incorrect. 
			& 40\% 
			\\
		    Empty SPARQL
			& The \texttt{WHERE}-clause of $\mathcal{S}_p$ does not have any relation triples or \texttt{FILTER}-clauses.
			& 22\% 
			\\
			\cmidrule(lr){1-3}
			Entity
			& The entities are filled into the wrong entity slots of the AQG when there are multiple entities in $\mathcal{S}_p$.
			& 4\%
			\\
			Relation
			& The incorrect relation instances are filled into the relation slots of the AQG.
			& 12\%
			\\
			Operator \& Value
			& Some operators and values in the \texttt{Filter}-clauses are incorrect.
			& 12\%
			\\
			\bottomrule
		\end{tabular}
		}
	\end{center}
\end{table*}

\begin{table} 
	\begin{center}
	{\caption{Online running time of an NLQ.\label{tab:time}}}
	\scalebox{0.95}{
		\begin{tabular}{lccccccccc}
			\toprule
			\multicolumn{1}{l}{\multirow{2}[1]{*}{\textbf{Method}}}
			&\multicolumn{2}{c}{\textbf{CWQ}}&\multicolumn{2}{c}{\textbf{LC-QuAD}}&\multicolumn{2}{c}{\textbf{WebQSP}}\\
			\cmidrule(lr){2-3} \cmidrule(lr){4-5} \cmidrule(lr){6-7}
			
			& $T_{\rm kg}$ &$T_{\rm inf.}$ 
			& $T_{\rm kg}$ &$T_{\rm inf.}$ 
			& $T_{\rm kg}$ &$T_{\rm inf.}$  \\
		    \cmidrule(lr){1-1} \cmidrule(lr){2-3} \cmidrule(lr){4-5} \cmidrule(lr){6-7}
			LR & Time Out &-
			&2,509 &513 
			&2,294 &457 \\
			StrLR & Time Out &- 
			&3,923 &395 
			&1,951 &350 \\
			\cmidrule(lr){1-1} \cmidrule(lr){2-3} \cmidrule(lr){4-5} \cmidrule(lr){6-7}
			HGNet &\textbf{5,030} &\textbf{576} 
			&\textbf{1,031} &\textbf{299}
			&\textbf{1,450} &\textbf{298} \\
			\bottomrule
		\end{tabular}
		}
	\end{center}
\end{table}

\subsubsection{Performance on Different Complexity and Types}

To further analyze the performance of the model on complex NLQs, we define the difficulty level of NLQs as the number of edges of the corresponding query graph because more edges lead to more semantic information. Fig. \ref{fig:level} shows the F1-scores on different levels. HGNet achieves the best results on almost all levels of NLQs. For the CWQ dataset, HGNet brings more significant improvements on NLQs above level 5 compared to \textsc{Bart} and NHGG. This again highlights the contribution of hierarchical graph generation to complex NLQs. Moreover, HGNet can compete with strong ranking-based methods at levels up to 4. We observe that the fluctuations are higher for each method on WebQSP because the number of high-level NLQs in this dataset is so small that any slight error can lead to a large gap. Interestingly, the improvement due to PLM seems to be independent of the difficulty level of the NLQs.

We also evaluated the performance of these methods when dealing with various SPARQL grammars, and the results are shown in Fig. \ref{fig:type}, where \texttt{Common}(Com.) denotes SPARQL with only relational triples.
Benefiting from the \textit{Outlining} process, the F1-scores of StrLR and HGNet are close on different syntax types for both LC-QuAD and WebQSP. However, on CWQ, HGNet has a clear advantage in handling {ORDER BY} clauses (Ord.), aggregation functions (Agg.), and nested queries, while both NHGG and {Bert} are powerless. Note that the improvement of \textsc{Bert} over Com. is more pronounced, which may prove that \textsc{Bert} is more suitable for identifying multi-hop relations than aggregation and comparison relations.

\subsubsection{Ablation Test}

To explore the contributions of each component of our HGNet, we compare the performance of the following settings:

    \noindent \textbf{$-$Graph Encoder} In \textit{Outlining}, we remove the graph encoder and modify Eq. (\ref{equ:graph_encoder}) to $\mathbf{h}_{in}^t = {\rm tanh}(W_{in} [\mathbf{h}_\mathcal{Q}^t; \mathbf{h}_{out}^{t-1}])$.
     
     \noindent \textbf{$-$Auxiliary Encoding} In \textit{Filling}, we remove the auxiliary structural encoding and modify Eq. (\ref{equ:auxiliary_encoding}) to $\mathbf{\hat{h}}_{in}^t = {\rm tanh}(\hat{W}^v_{in} \mathbf{\hat{h}}_\mathcal{Q}^t)$.
     
     \noindent \textbf{$-$Copy} During \textit{Filling} we remove the copy mechanism and modify Eq. (\ref{equ:iv}) to $i_v^t = \hat{i}_v^t$.
     
     \noindent \textbf{$-$EG} We remove the execution guidance strategy thereby performing edge-\textit{Filling} only by the model prediction.
     
     \noindent \textbf{Repl. LSTM with Transformer} We replace the LSTM with a randomly initialized 6-layer Transformer~\cite{DBLP:conf/nips/VaswaniSPUJGKP17} (encoder-decoder) as the backbone of our HGNet.
    
    Table \ref{tab:ablation_test} shows the $\mathcal{G}_a$ Acc, $\mathcal{G}_q$ Acc and F1 scores for different settings. By removing the graph encoder, $\mathcal{G}_a$ Acc decreases by an average of 7\% on all datasets. This proves that the full structural information of the graph in the previous step is crucial for \textit{Outlining}. Discarding the auxiliary structural encoding leads to a decrease in $\mathcal{G}_q$ Acc for all datasets (-2.5\%, -0.9\%, -4.0\%), while there is a small increase in $\mathcal{G}_a$ Acc for LC-QuAD (0.1\%). One possible reason is that the auxiliary encoding may affect the original information flow of the structure prediction. In contrast, the copy mechanism plays an important role in the $\mathcal{G}_q$ Acc of LC-QuAD, since a considerable number of SPARQLs have duplicate relations. On all datasets, the removal of EG brought an absolute decrease in Acc (-7.5\%, -28.6\%, -5.3\%). This proves that KG is effective for the disambiguation of query graphs. EG contributes less to CWQ and WebQSP than LC-QuAD, which is based on DBpedia and has \texttt{property} and \texttt{ontology} conflicts. The Transformer does not significantly improve the performance compared to LSTM and takes a longer time to infer. This is probably because in each decoding step, it needs to take all previous steps as input.

\subsubsection{Few-shot Performance}
To explore the odds performance of HGNet, we trained the model with different sizes of training data (CWQ for {1\%, 5\%, 10\%, 30\%} and LC-QuAD and WebQSP for {5\%, 10\%, 30\%, 50\%}). The results are shown in Fig. \ref{fig:few_shot}. For each size of CWQ, the full HGNet equipped with all components maintained the best performance and showed a consistent improvement over the other ablation settings. The main reason for its poor performance on the 5\% setting of LC-QuAD is that the training data are too small (only 175 NLQs) for the model to learn to generate both structural and replicated instances. However, this is not the case with WebQSP, as most of its SPARQL procedures are simple to learn.

\subsubsection{Error Analysis}
To understand the sources of errors, we analyzed 100 random HGNet failure examples on the CWQ test set. We summarized several main reasons for the failures, see Table \ref{tab:errors}. The top shows the errors in structure prediction, and the bottom shows the errors in instance prediction. Note that the sum of all proportions exceeds 100\%, because a sample may have multiple errors at the same time. For example, a result query graph has an edge with the wrong orientation (structure error) and also has a vertex with an incorrect entity instance (instance error).
The structure errors are mainly reflected in the redundancy, missingness, and incorrect topology of the relation triples. For example, for the NLQ ``The Jade Emperor deity is found in what sacred text(s)", the correct SPARQL condition should be described as \{?c \textit{deities} Jade\_Emperor, ?c \textit{texts} ?x\}, but our result is \{?x \textit{deities} Jade\_Emperor\}.  
Sometimes, HGNet drops all running beams due to conflicts between the prediction structure and candidate instances in the KG, resulting in empty SPARQLs. For example, suppose a beam produces an incorrect AQG $\mathcal{G}_a$ and the model currently needs to predict instances of the relational edge $e$ in $\mathcal{G}_a$.
With the EG strategy, it finds that only \textit{character} is a legitimate instance, but the candidate instance pool $\mathcal{I}_{rel}$ does not contain \textit{character}. The beam cannot continue and is therefore discarded. When all the beams are discarded, the generation ends with an empty query graph.
The current bottleneck of HGNet is still the semantic understanding of NLQ. Some instances are not mentioned in any explicit way and can only be identified by external knowledge. For example, the NLQ ``What is lawton ok" is similar in form to ``What is f. scott fitzgerald", but corresponds to two different relations, \textit{postal\_codes} and \textit{profession}, respectively. This problem is improved to some extent when we enhance HGNet with the pre-trained \textsc{Bert}.

\subsubsection{Efficiency Evaluation}
We also compared the online running times of our HGNet with those of our two main competitors using three test sets. To evaluate the efficiency more rationally, for an NLQ, we split the running time of each method into two parts: (1) $T_{\rm kg}$ represents the interaction time with the KG using SPARQLs, such as LR and our EG's candidate query graph collection; (2) $T_{\rm inf.}$ represents the model inference time, which is affected by the number of model parameters and the size of the search space. We set an upper limit of $100,000$ ms for the execution time of each SPARQL. The results are shown in Table \ref{tab:time}. 
LR and StrLR do not have $T_{\rm inf.}$ results in CWQ, which is attributed to the huge search space that prevents the neural network from running. Our HGNet achieves the highest efficiency on both $T_{\rm kg}$ and $T_{\rm inf.}$. For $T_{\rm kg}$, although our EG strategy has more interactions, it relies on \texttt{ASK} intent, which is far faster than the \texttt{SELECT} intent collected by the query graph. For $T_{\rm inf.}$, HGNet benefits from a significant reduction in the search space, while its larger parameter size makes the efficiency improvement not as significant as expected in Section \ref{sec:search_space}.

\section{Related Work}
\textbf{Knowledge Graph Question Answering}
The goal of KGQA~\cite{DBLP:conf/acl/KapanipathiARRG21,DBLP:conf/semweb/HuSHQ21} is to provide clear answers for NLQs. In general, the solutions can be divided into two main categories. 
The first category is based on semantic parsing (SP), which translates NLQs into logical form. Our HGNet belongs to this category. Earlier works dealt with this problem by parsing or predefined templates~\cite{DBLP:conf/emnlp/BerantCFL13,DBLP:conf/acl/BerantL14}. With the development of advanced deep learning, later methods~\cite{DBLP:conf/acl/YihCHG15,DBLP:conf/acl/YuYHSXZ17,DBLP:journals/tkde/Hu0YWZ18,DBLP:conf/emnlp/Hu0Z18,DBLP:conf/emnlp/DingHXQ19,DBLP:journals/kbs/ChenL20} establish complex pipelines through neural networks.
Yih et al.\cite{DBLP:conf/acl/YihCHG15} proposed a staged query graph generation (STAGG) strategy, which utilizes a convolutional neural network (CNN) to detect the relation chains. 
% Hu et al.\cite{DBLP:conf/emnlp/Hu0Z18} define four operations to perform a state-transition generation process to build query graphs. Different from our proposed operations, theirs rely on the manually defined rules to identify the conditions on whether to perform the operation. Recent methods simplify pipeline into query graph ranking~\cite{DBLP:conf/emnlp/LuoLLZ18,DBLP:conf/semweb/MaheshwariTLCF019,DBLP:conf/ijcai/ChenLHQ20}. 
Hu et al.~\cite{DBLP:conf/emnlp/Hu0Z18} defined four operations to perform the state transition generation process to generate the query graph. Their operations rely on predefined conditions to determine when and where to execute the operations~\cite{DBLP:conf/emnlp/LuoLLZ18,DBLP:conf/semweb/MaheshwariTLCF019,DBLP:conf/ijcai/ChenLHQ20}.
Recent methods simplify the pipeline to learn to rank query graphs.
Luo et al.~\cite{DBLP:conf/emnlp/LuoLLZ18} applied the STAGG strategy to generate candidate query graphs and then ranked them by using a semantic matching model.
Maheshwari et al.~\cite{DBLP:conf/semweb/MaheshwariTLCF019} also follow the STAGG strategy and propose a self-attention-based slot matching model.
The major difference between our method and previous methods is that we propose a hierarchical generation framework to leverage the structural constraint to narrow the search space. Furthermore, our method does not need predefined templates and conditions. 
The second category is based on information retrieval (IR), which does not produce logical forms and focuses on scoring candidate answers directly through end-to-end models, such as key-value memory networks and graph convolutional networks~\cite{DBLP:conf/acl/YaoD14,DBLP:conf/emnlp/BordesCW14,DBLP:conf/acl/DongWZX15,DBLP:conf/emnlp/MillerFDKBW16,DBLP:conf/emnlp/SunDZMSC18,DBLP:conf/aaai/ZhangDKSS18,DBLP:conf/emnlp/SunBC19}. 

% The third category is based on Multi-hop Reasoning (MR), which use some classical reasoning models, such as key-value memory networks and graph convolutional networks~\cite{DBLP:conf/emnlp/MillerFDKBW16,DBLP:conf/emnlp/SunDZMSC18,DBLP:conf/aaai/ZhangDKSS18,DBLP:conf/emnlp/SunBC19}.

\noindent \textbf{Intermediate Representations in KGQA}
The traditional methods~\cite{DBLP:conf/emnlp/BerantCFL13,DBLP:conf/acl/BerantL14} utilize $\lambda$-DCS, which can be viewed as a syntactic tree, as an intermediate representation (IR) to represent the meaning of the question. Yih et al.~\cite{DBLP:conf/acl/YihCHG15} proposed representing  SPARQL with a query graph, which consists of subject entities, core relation chains and constraint nodes. This was the first time that the logical form of the representation was treated from a graph perspective. However, their query graphs are too coarse for the more complicated SPARQL syntax. For example, nested structures cannot be represented by a single constraint node. In contrast, our redefined query graph treats the SPARQL items of entities, values, types, relations, and built-in properties as real vertices or edges, and thus has a stronger representation capability.

\noindent \textbf{Grammar-based Decoding} The goal of grammar-based decoding is to generate a top-down grammar rule at each step of autoregressive decoding and has received increasing attention in text-to-SQL~\cite{DBLP:conf/emnlp/YuYYZWLR18,DBLP:conf/acl/GuoZGXLLZ19,DBLP:conf/acl/WangSLPR20}, which is another semantic parsing task. Our generation process depends on graph grammar rather than top-down grammar. It is typically difficult for the model to directly forecast (outline) the entire structure from the root node, while our method is not limited by tree hierarchy and thus could add new partial semantic information to the graph at any time of \textit{Outlining}.

\section{Conclusion}
In this paper, we present a new method for complex knowledge graph question answering (KGQA) that transforms NLQs into executable query graphs. We initially redefine the grammar of query graphs to represent complicated SPARQL syntax and propose a new hierarchical query graph generation model. It completes the query graph by first outlining the structure of the query graph and then populating the structure with instances. 
Benefiting from the structural constraints provided by the hierarchical generation procedure, our model greatly reduces the search space for candidates while avoiding local ambiguity. Experimental results demonstrate that our model achieves better results than existing methods, especially on complex NLQs. In future work, we will try to enhance the model with pre-trained language models to exploit the knowledge of a large corpus. We will also try to port the model to a weakly supervised condition with only NLQ-answer pairs.

% or


% \appendices

% % you can choose not to have a title for an appendix
% % if you want by leaving the argument blank
% \section{}

% % use section* for acknowledgment
% \ifCLASSOPTIONcompsoc
%   % The Computer Society usually uses the plural form
%   \section*{Acknowledgments}
% \else
%   % regular IEEE prefers the singular form
  \section*{Acknowledgment}
  This work is supported by the Natural Science Foundation of China (Grant No. 61502095, U21A20488).
% \fi


\ifCLASSOPTIONcaptionsoff
  \newpage
\fi



\bibliographystyle{IEEEtran}
\bibliography{main}
\newpage


% \appendix[Proof of the Zonklar Equations]
% or
% \appendix  % for no appendix heading


\appendices
\section{Preprocessing of Transforming SPARQL to Query Graph\label{app:pre-process_sparql}}
\subsection{Constraint of Time Interval\label{app:temporal_constraint}}
\begin{figure*}[!t]
\centering
\includegraphics[width=0.9\textwidth]{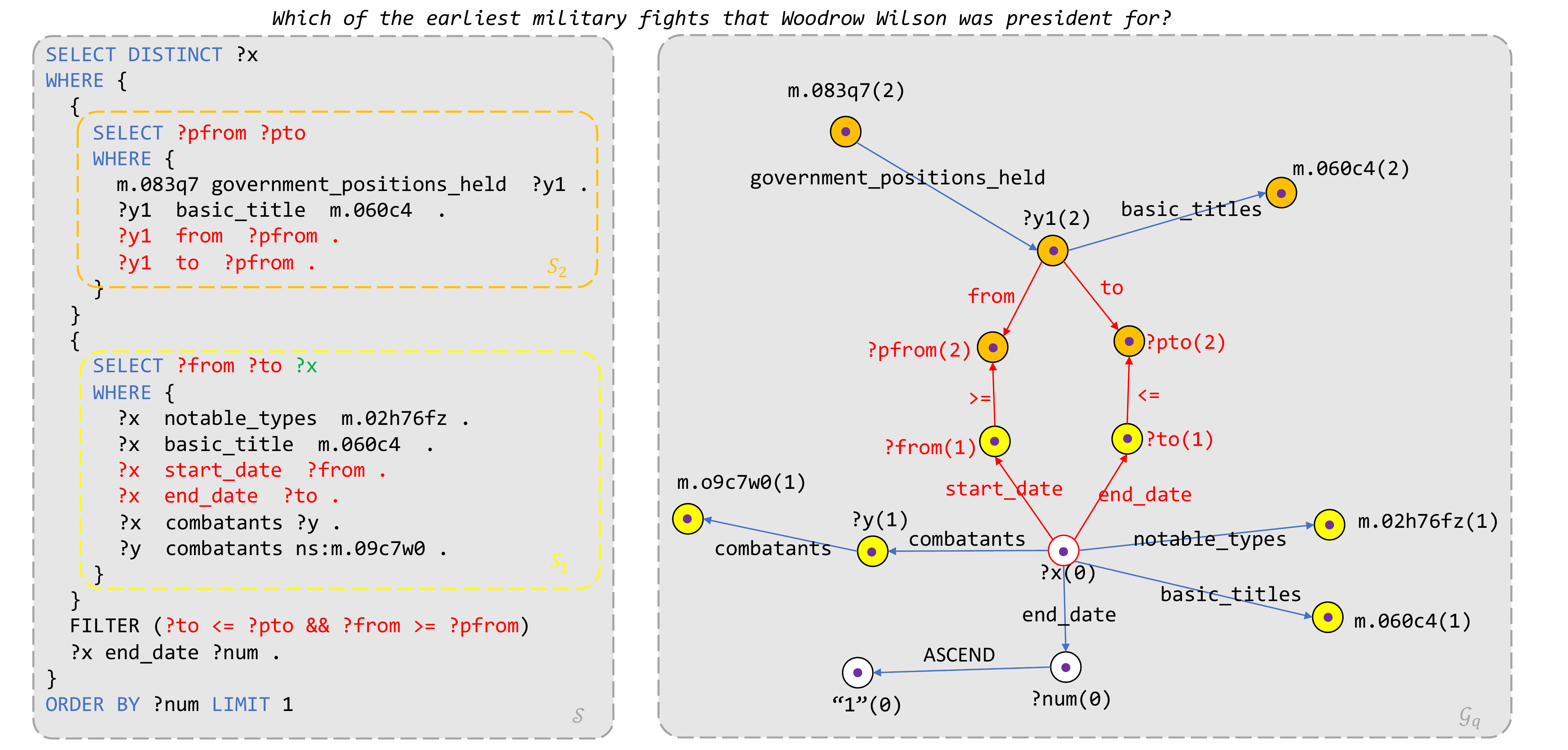}
\caption{An example SPARQL program $\mathcal{S}$ having the constraint of time interval and the subquery with \texttt{?x} intention.}
\label{fig:more_complex_query_graph}
\end{figure*}

We observe that some SPARQL programs have subqueries that provide time intervals as constraints. Fig. \ref{fig:more_complex_query_graph}a shows such a SPARQL program $\mathcal{S}$ in CWQ. It has two subqueries, namely $\mathcal{S}_1$ (yellow box) and $\mathcal{S}_2$ (orange box), which aim to query time intervals \texttt{[?from, ?to]} and \texttt{[?pfrom, ?pto]}. Here, \texttt{?from} and \texttt{?to} respectively denote the start time and end time of being \textit{President} (\texttt{m.060c4}) for \textit{Woodrow Wilson} (\texttt{m.083q7}). \texttt{?from} and \texttt{?to} respectively denote the start time and end time of the \textit{Military Conflict} (\texttt{m.02h76fz}). $\mathcal{S}_1$ and $\mathcal{S}_2$ are connected by a \texttt{FILTER} clause (green box). The clause describes the constraint that period \texttt{[?from, ?to]} is during period \texttt{[?pfrom, ?pto]}. 

% Such constraints of time intervals typically result in a problem: the corresponding query graph $\mathcal{G}_q$ does not follow Property \ref{prop:vertex_number} because it has the following generalized cycle path (red path in Fig. \ref{fig:more_complex_query_graph}) without considering the direction, thereby can not be handled by our generation framework. 
Unfortunately, such constraints with time intervals prevent the corresponding query graph $\mathcal{G}_q$ from being processed by our generative framework.
This is because it has the following generalized cycle (red path in Fig. \ref{fig:more_complex_query_graph}) and thus does not satisfy Property \ref{prop:vertex_number}.

\begin{equation}
\begin{aligned}
{\rm ?x(0)} &{\rm \stackrel{start\_date}{\longrightarrow}   ?from(1) \stackrel{\ge}{\longrightarrow}   ?pfrom(2) \stackrel{from}{\longleftarrow}   ?y1(2)} \\ 
&{\rm \stackrel{to}{\longrightarrow}   ?pto(2) \stackrel{\le}{\longleftarrow}   ?to(1)
\stackrel{end\_date}{\longleftarrow}   ?x(0)}
\end{aligned}
\nonumber
\end{equation}

We propose a simple but efficient strategy to avoid the cycle produced by time intervals. Initially, we retrieve all the time intervals in the SPARQL program by checking the \texttt{FILTER} clauses because a time variables are always in \texttt{FILTER} clauses and connected with the described variable (e.g., \texttt{?x}) along a relation $r$ (e.g., \texttt{start\_date}). We let $p=[t_{st}$, $t_{ed}]$ denote the time interval, and $r_{st}$ and $r_{ed}$ denote the two relations connected to $t_{st}$ and $t_{ed}$, respectively. They are combined into a new relation $r_p$. In addition, to describe the comparison relationship between time intervals, we define the following two new operators:
\begin{itemize}
    \item \texttt{DURING}: $\forall p^1=[t_{st}^1, t_{ed}^1], \forall p^2=[t_{st}^2, t_{ed}^2]$, $p^1\quad \texttt{DURING} \quad p^2$ represents  $t_{st}^1 \ge t_{st}^2 \quad \texttt{AND} \quad t_{ed}^1 \le t_{ed}^2$.
    \item \texttt{OVERLAP}: $\forall p^1=[t_{st}^1, t_{ed}^1], \forall p^2=[t_{st}^2, t_{ed}^2]$, $p^1\quad \texttt{OVERLAP} \quad p^2$ represents  $t_{st}^1 \le t_{ed}^2 \quad \texttt{AND} \quad t_{ed}^1 \ge t_{st}^2$.
\end{itemize}
In this way, the cycle in Fig. \ref{fig:more_complex_query_graph} is converted to the following chain.
\begin{equation}
\begin{aligned}
{\rm ?x(0)} {\rm \stackrel{start\_date\$\$\$end\_date}{\longrightarrow}   ?p(1) \stackrel{\texttt{DURING}}{\longrightarrow}   ?p1(2) \stackrel{from\$\$\$to}{\longleftarrow}   ?y1(2)}
\end{aligned}
\nonumber
\end{equation}
where both ?p(1) and ?p1(2) are the time intervals, and from\$\$\$to and start\_date\$\$\$end\_date are the combined relations. Fig. \ref{fig:temporal_interval} shows $\mathcal{S}$ and the new query graph after combining. By the observation, we found that only two kinds of relation pairs are used to describe time intervals, namely (x.from, x.to) and  (x.start\_date, x.end\_date), where x is the domain prefix. Therefore, we collect all such relations to combine each pair of them into a new relation. In the first stage of our proposed method, these new relations are also added to the relation pool with the normal relations, which are all fed to the relation ranker to select candidate instances. During inference, HGNet could fill them into the edge slots to construct the query graph. Finally, when the completed query graph is converted to the SPARQL program, such combined relations are split into the pair of normal relations to restore the cycle path by simple inverse rules. 

\subsection{Sub-query with ?x Intention}
\begin{figure*}[!t]
\centering
\includegraphics[width=0.9\textwidth]{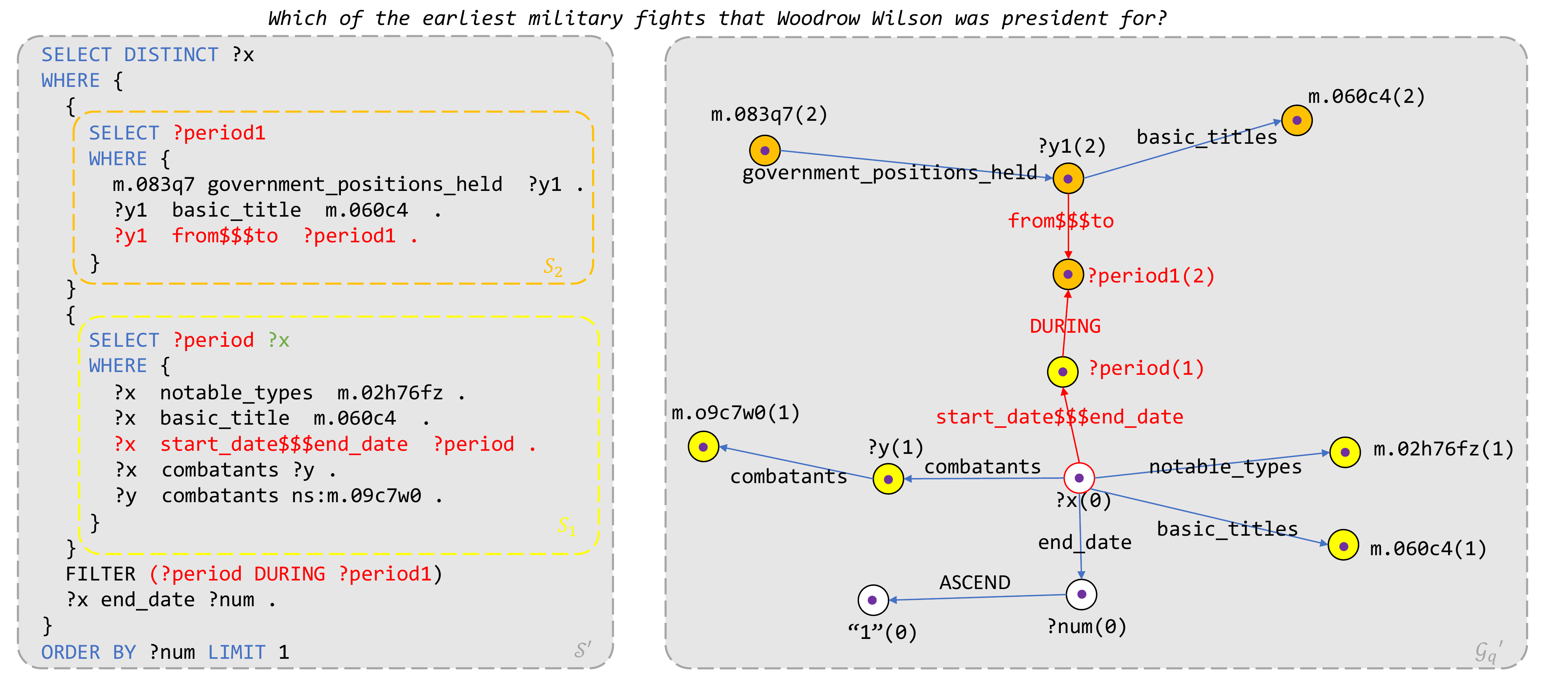}
\caption{$\mathcal{S}'$ and $\mathcal{G}_q'$ obtained by removing the cycle path of time intervals.}
\label{fig:temporal_interval}
\end{figure*}

The SPARQL program $\mathcal{S}$ in Fig. \ref{fig:temporal_interval} still has a problem where answer \texttt{?x} not only appears in the \texttt{SELECT}-clauses of the main query but also appears in those of subquery $\mathcal{S}_1$ (yellow). The problem makes the segment number of \texttt{?x} hard to determine. In $\mathcal{G}_q$, segment 1 (yellow) corresponding to $\mathcal{S}_1$ is divided into multiple parts by \texttt{?x(0)} (white) belonging to segment 0. Although it does not influence the generation process of HGNet, it is difficult to restore the final generated query graph to $\mathcal{S}$ due to the broken segment.

\begin{figure*}[!t]
\centering
\includegraphics[width=0.9\textwidth]{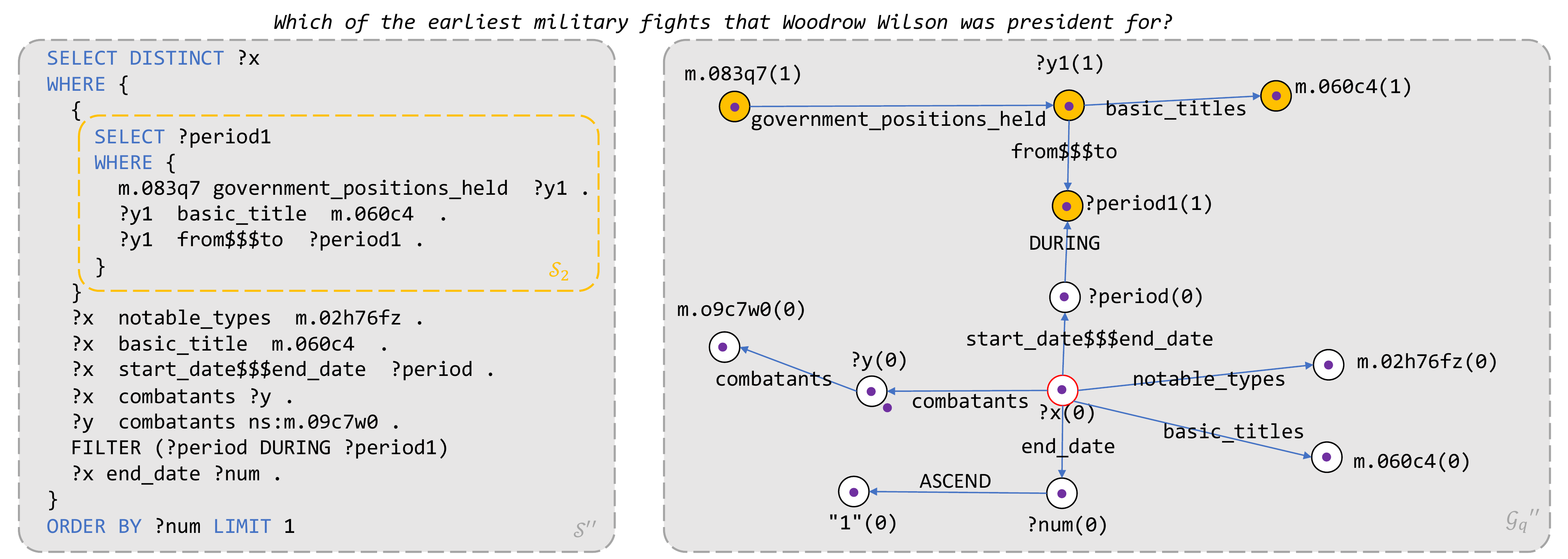}
\caption{$\mathcal{S}''$ and $\mathcal{G}_q''$ obtained by simplifying \texttt{?x} intention, which are used for training HGNet.}
\label{fig:subquery_x_intention}
\end{figure*}

To solve this problem, we propose to merging such a sub-query with \texttt{?x} intention (i.e., $\mathcal{S}_2$) with the main query before transforming SPARQL to the query graph. The obtained new SPARQL $\hat{\mathcal{S}}$ is shown in Fig. \ref{fig:subquery_x_intention}. In fact,  $\mathcal{S}$ and $\hat{\mathcal{S}}$ are semantically equivalent and have identical retrieved answers, while the query graph of the latter retains complete segments so that they can be easily restored to an executable SPARQL program.

\subsection{EXISTS Clause in FILTER Clause}
\begin{figure}[t]
\centering
\includegraphics[width=0.5\textwidth]{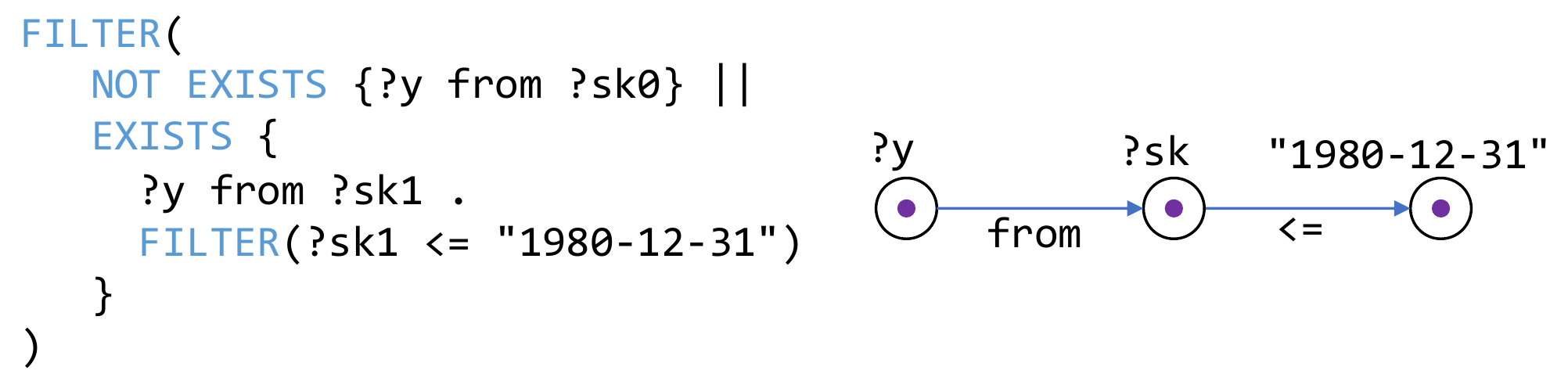}
\caption{An example of an \texttt{EXISTS} clause in the \texttt{FILTER} clause.}
\label{fig:exist_in_filter}
\end{figure}

In our used datasets, the \texttt{EXISTS} clause usually appears in the \texttt{FILTER} clause. Considering the example on the left of Fig. \ref{fig:exist_in_filter}, the SPARQL subprogram describes a constraint $\mathcal{C}$ that the start time of \texttt{?y} is less than \texttt{1980-12-31}. Here, \texttt{EXISTS} aims to represent the case in which \texttt{?y} has the start time and \texttt{NOT EXISTS} aims to address that \texttt{?y} does not have the start time. $||$ is used for the conjunction of the two cases above to prevent execution failure of the SPARQL. Actually, the two \texttt{EXISTS} clauses are not associated with the semantic information of the NLQ but are merely the underlying implementation of SPARQL and the KG. Therefore, as the intermediate representation, our query graph should not represent such \texttt{EXISTS} clauses to eliminate the gap between the NLQ and SPARQL.  The query graph needs only to describe the constraint $\mathcal{C}$, which is shown on the right of Fig. \ref{fig:exist_in_filter}.

\section{More Examples of Query Graphs}
\begin{figure*}
  \centering
    \begin{subfigure}{0.48\textwidth}
      \centering   
      \includegraphics[width=\linewidth]{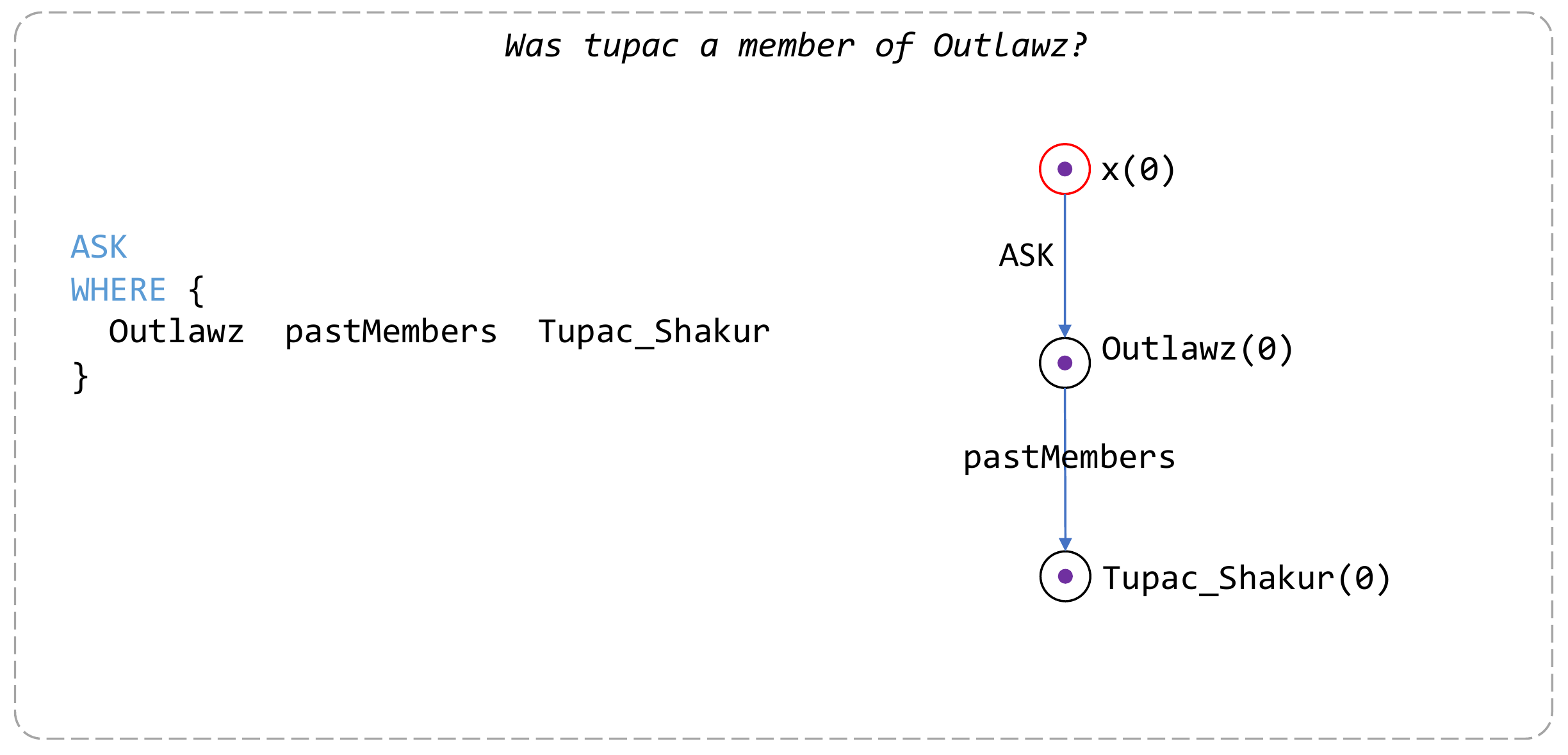}
        \caption{}
    \end{subfigure}   %      \hfill  %
    \begin{subfigure}{0.48\textwidth}
      \centering   
      \includegraphics[width=\linewidth]{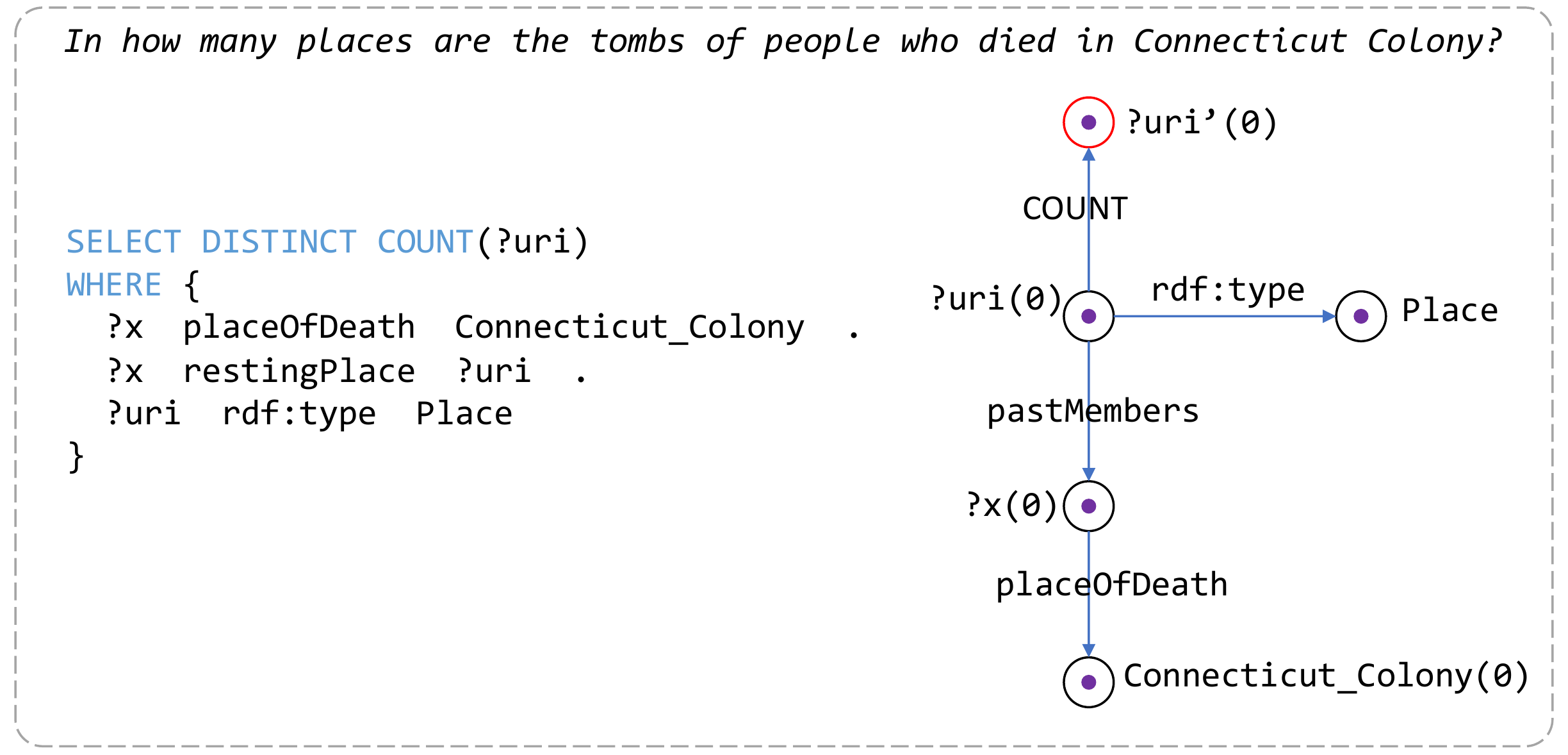}
        \caption{}
    \end{subfigure}
    \begin{subfigure}{0.48\textwidth}
      \centering   
      \includegraphics[width=\linewidth]{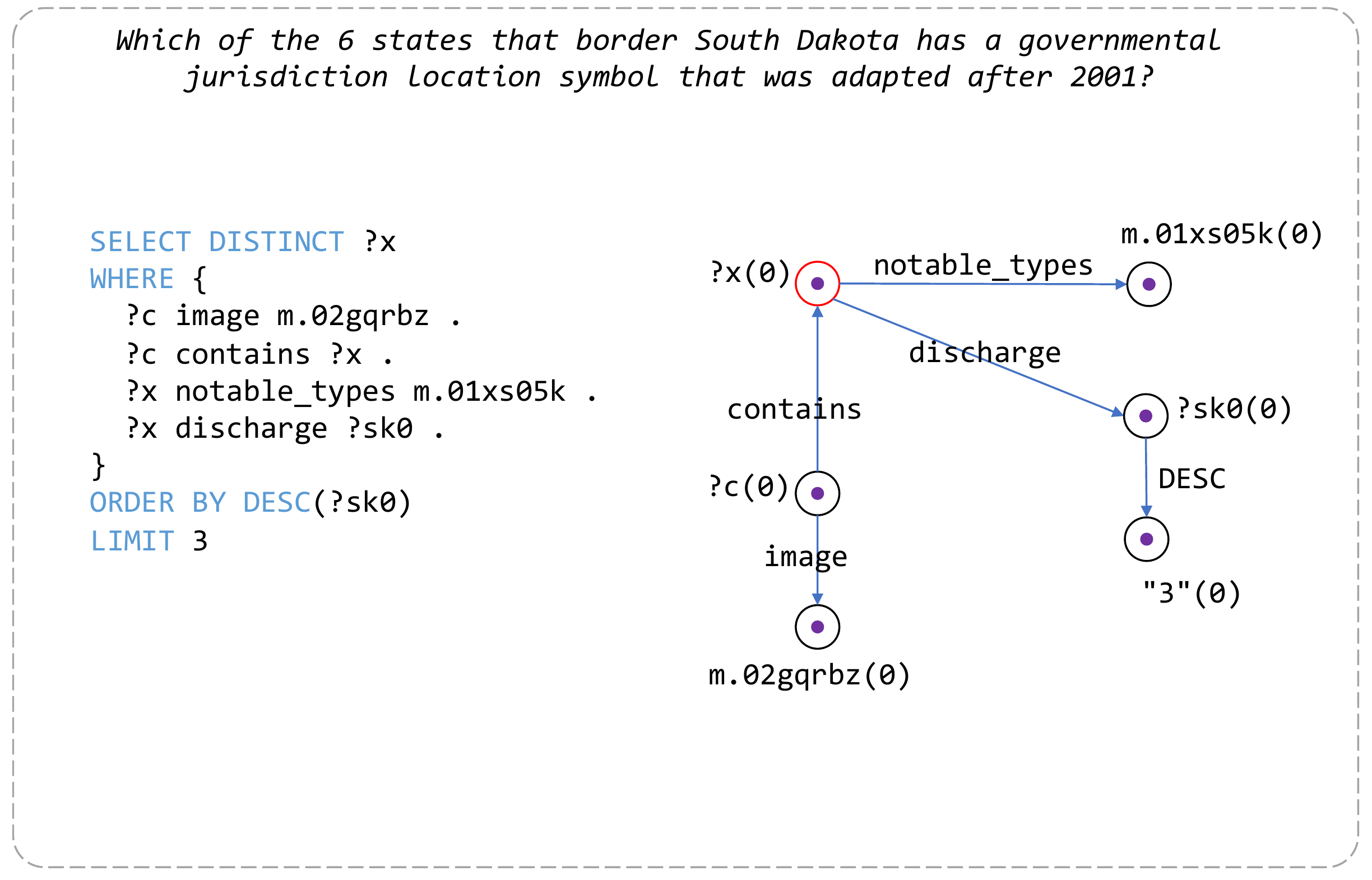}
        \caption{}
    \end{subfigure}   %      \hfill  %
    \begin{subfigure}{0.48\textwidth}
      \centering   
      \includegraphics[width=\linewidth]{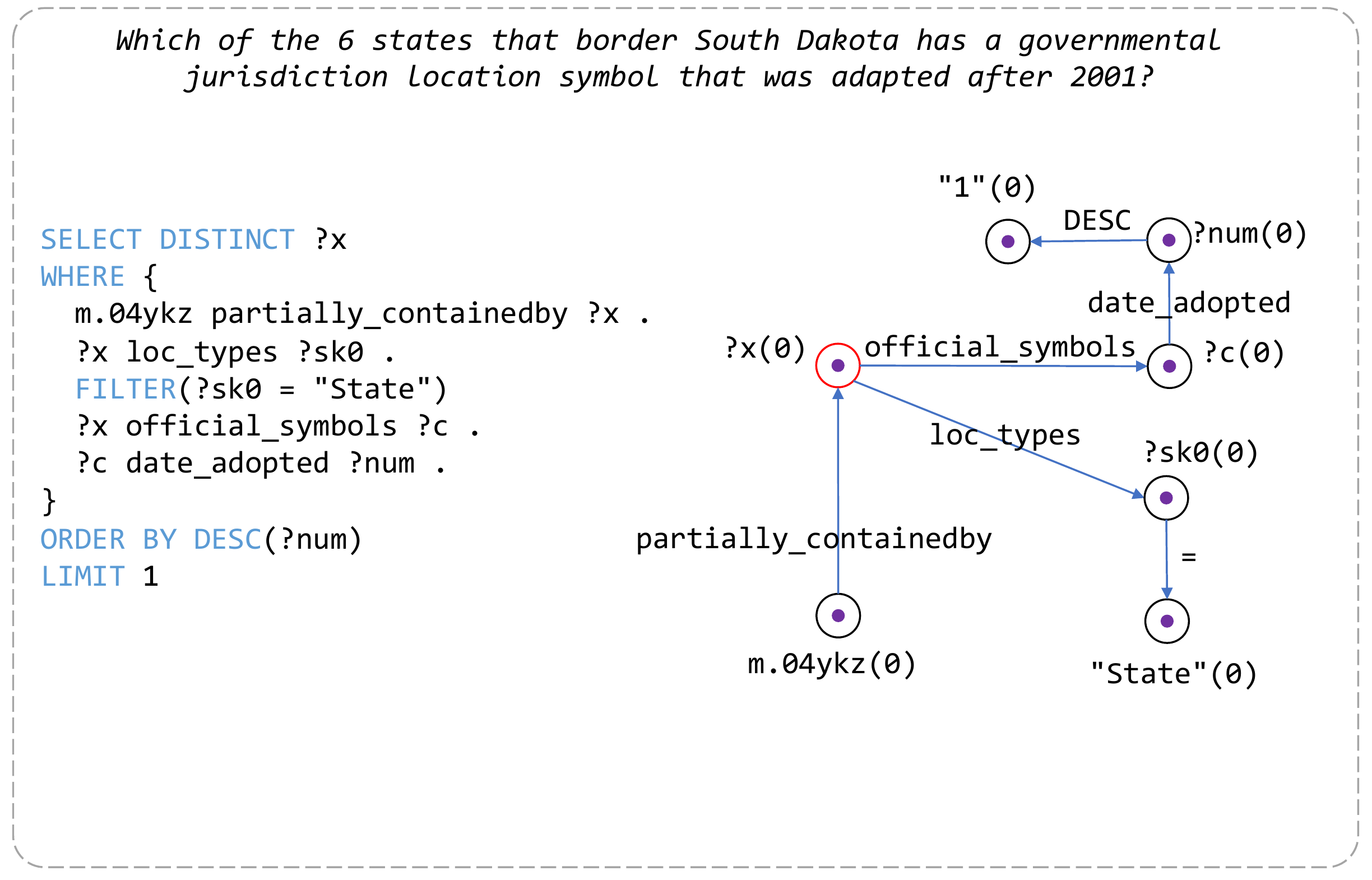}
        \caption{}
    \end{subfigure}
    \begin{subfigure}{0.48\textwidth}
      \centering   
      \includegraphics[width=\linewidth]{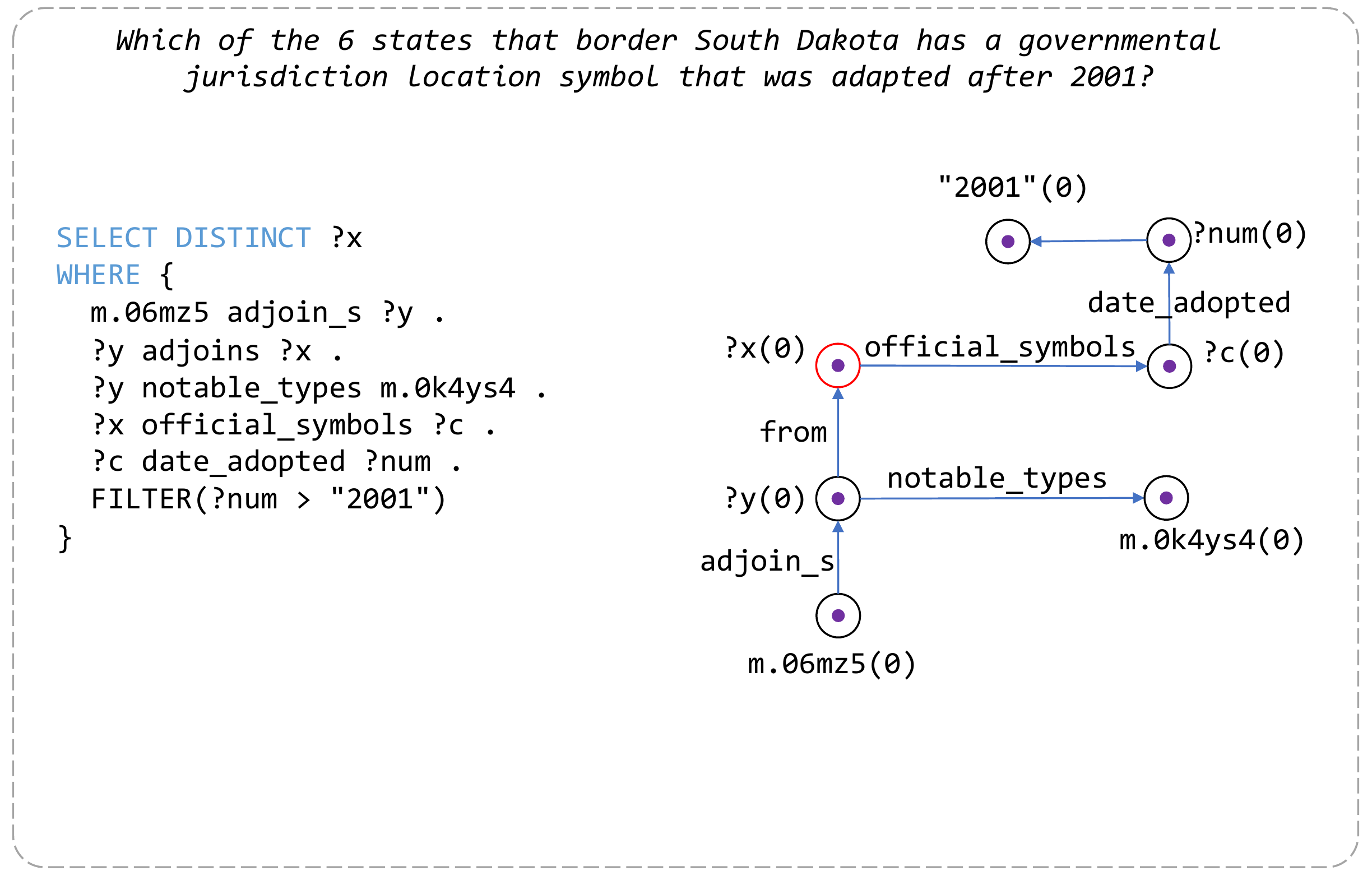}
        \caption{}
    \end{subfigure}   %      \hfill  %
    \begin{subfigure}{0.48\textwidth}
      \centering   
      \includegraphics[width=\linewidth]{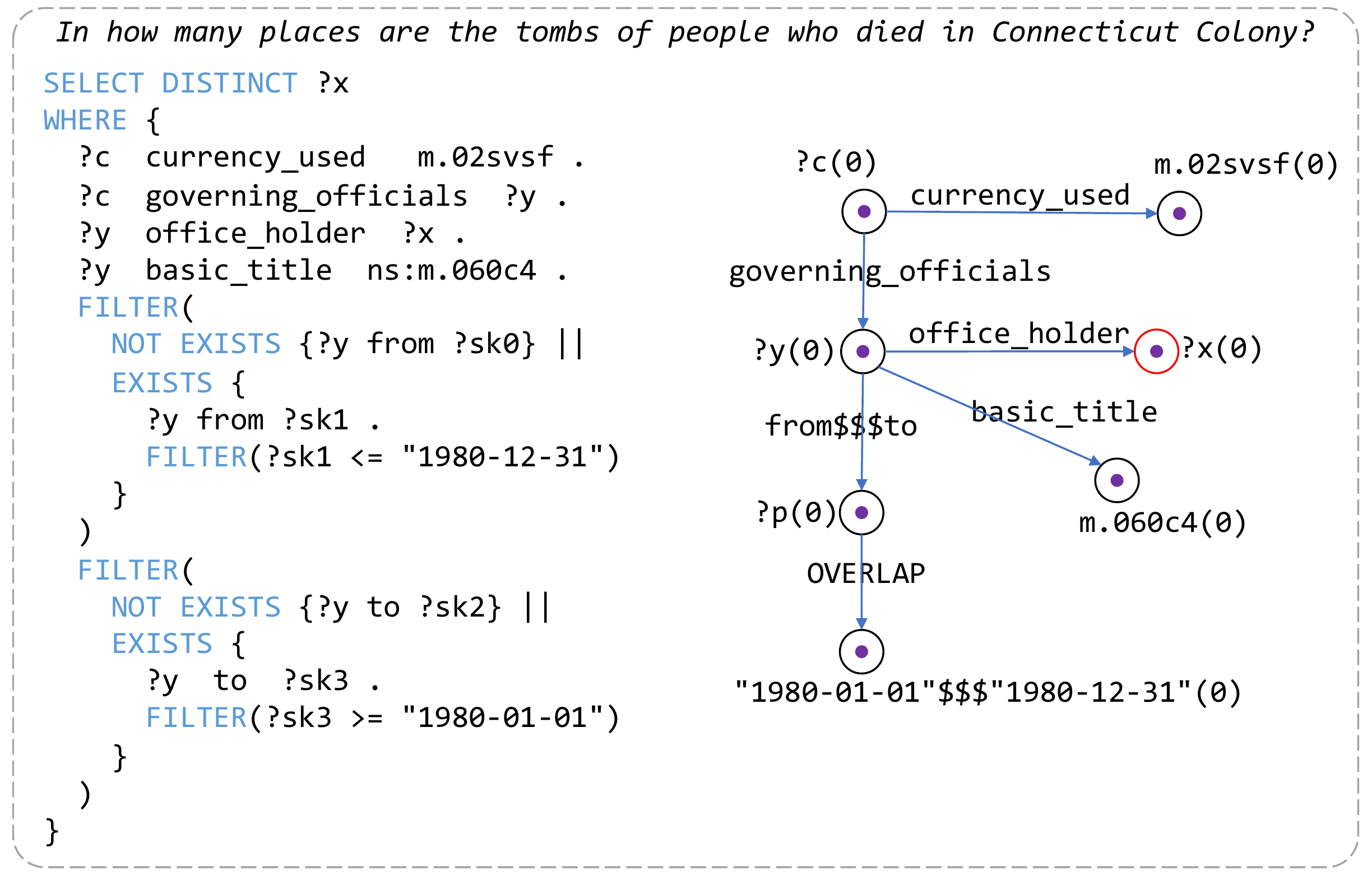}
        \caption{}
    \end{subfigure}
    \begin{subfigure}{0.48\textwidth}
      \centering   
      \includegraphics[width=\linewidth]{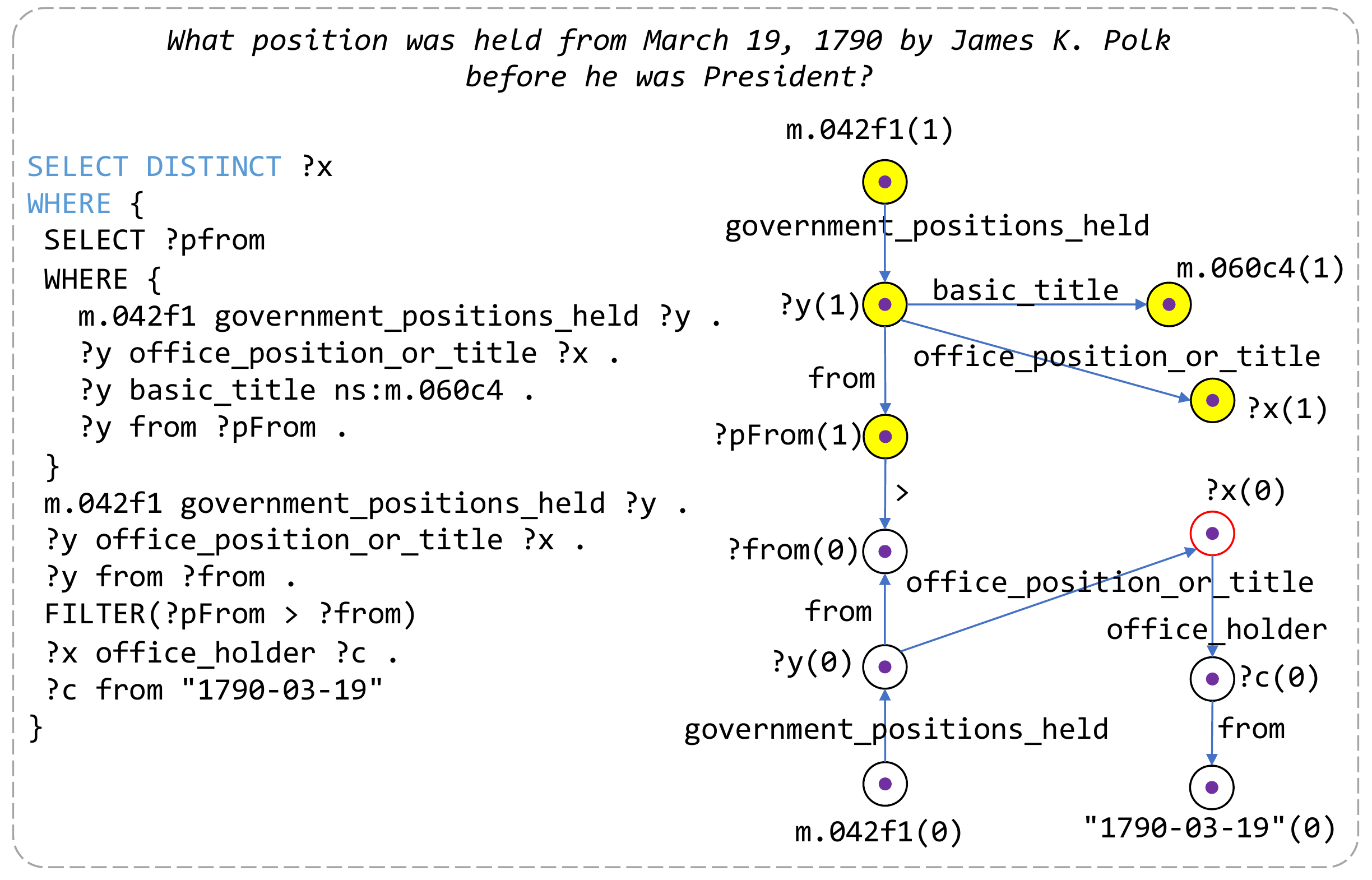}
        \caption{}
    \end{subfigure}   %      \hfill  %
    \begin{subfigure}{0.48\textwidth}
      \centering   
      \includegraphics[width=\linewidth]{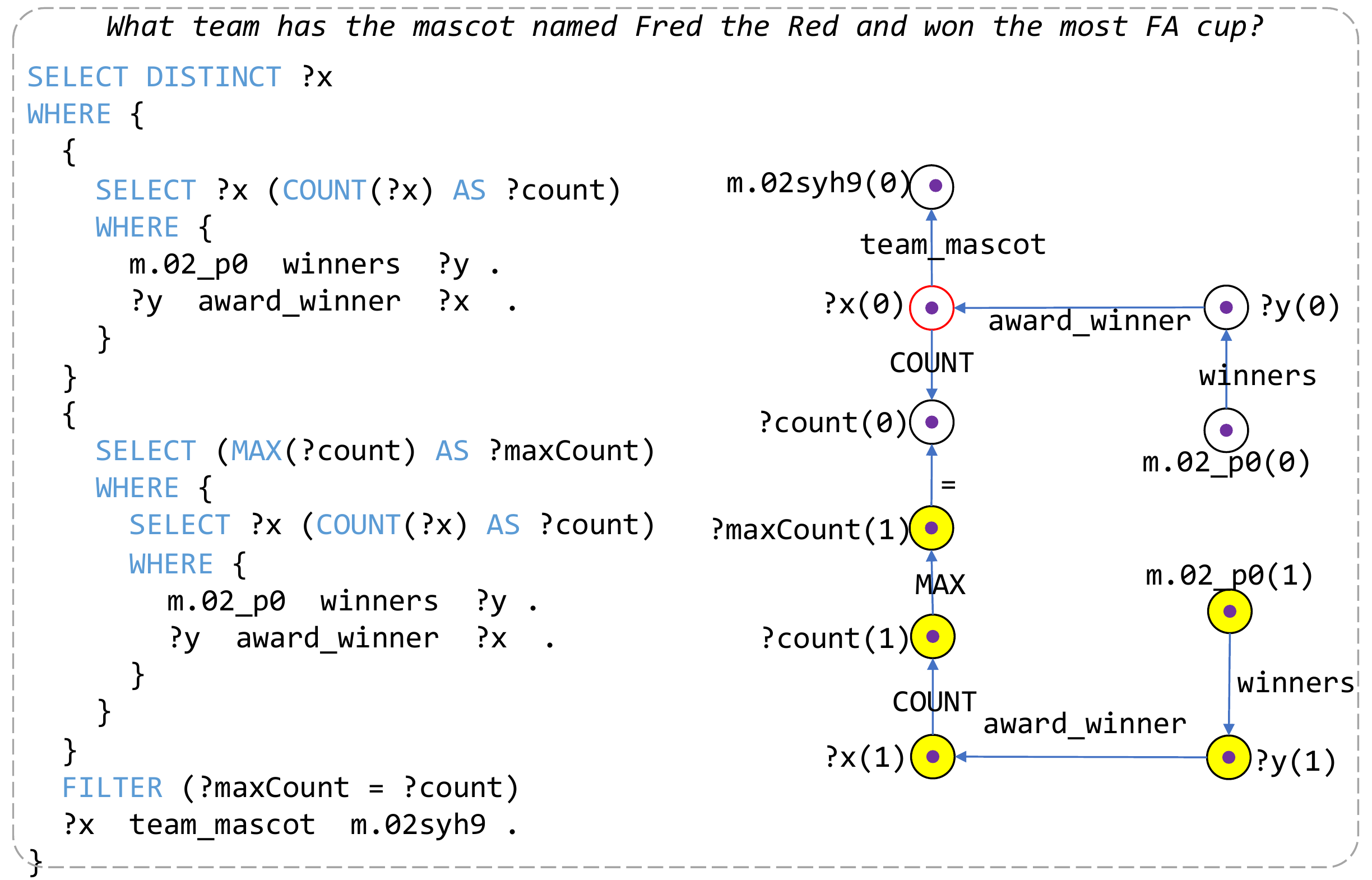}
        \caption{}
    \end{subfigure}
\caption{
\label{fig:more_examples}
More examples of our redefined query graphs. Red denotes the answer vertex.
}
\end{figure*}

Fig. \ref{fig:more_examples} presents more examples of our redefined query graphs.

\section{Supervised Signal Construction for HGNet}
\label{sec:signal_building}
The supervised signals for training HGNet, i.e., $\mathcal{A}_o$, $\mathcal{A}_v$, and $\mathcal{A}_e$, are obtained in Algorithm \ref{alg:signal_building}. 

\begin{algorithm}[t]
\caption{Supervised Signal Construction for HGNet \label{alg:signal_building}}
\begin{algorithmic}[1]
\Require A query graph $\mathcal{G}_q = (V_q, E_q, \Psi_q, \Phi_q, S_q)$.
\State Initialize \textit{Outlining} supervised signals $\mathcal{A}_o=[]$, vertex \textit{Filling} supervised signals $\mathcal{A}_v=[]$, edge \textit{Filling} supervised signals $\mathcal{A}_e=[]$.

\Function{DepthFirstTraversal}{$u$}
\For{vertex $v \in V_q - \{u\}$}
\State Set edge $e^+ = \left \langle u, v \right \rangle$, $e^- = \left \langle v, u \right \rangle$,
%\State Set segment increment $s_{\delta} = s_v - s_u$
\State Set copied vertex $\hat{v} = \textsc{GetCopy}(v)$
\State Set copied edge $\hat{e}^+ = \textsc{GetCopy}(e^+)$, $\hat{e}^- = \textsc{GetCopy}(e^-)$
\State Set vertex instance $i_v = \textsc{GetInstance}(v)$
\State Set edge instance $i_{e^+} = \textsc{GetInstance}(e^+)$, $i_{e^-} = \textsc{GetInstance}(e^-)$
\If{$e^+ \in E_q$}
\State $\mathcal{A}_o$.\textsc{Append}($[c_v, s_{v}, \hat{v}]$) \label{algline:outlining_st}
\State $\mathcal{A}_o$.\textsc{Append}($[u]$)
\State $\mathcal{A}_o$.\textsc{Append}($[c_{e^+}, \hat{e}^+]$) \label{algline:outlining_ed}
\State $\mathcal{A}_v$.\textsc{Append}($i_v$) \label{algline:filling_st}
\State $\mathcal{A}_e$.\textsc{Append}($i_{e^+}$) \label{algline:filling_ed}
\State \Call{DepthFirstTraversal}{$v$}
\ElsIf{$e^- \in E_q$}
\State $\mathcal{A}_o$.\textsc{Append}($[c_v, s_{v}, \hat{v}]$) \label{algline:incoming_st}
\State $\mathcal{A}_o$.\textsc{Append}($[u]$)
\State $\mathcal{A}_o$.\textsc{Append}($[c_{e^-}, \hat{e}^-]$)
\State $\mathcal{A}_v$.\textsc{Append}($i_v$)
\State $\mathcal{A}_e$.\textsc{Append}($i_{e^-}$) \label{algline:incoming_ed}
\State \Call{DepthFirstTraversal}{$v$}
\EndIf
\EndFor
\EndFunction
\State $\mathcal{A}_o$.\textsc{Append}($[\texttt{Ans}, 0, \texttt{NONE}]$)\label{alqline:first_operation_st}
\State $\mathcal{A}_v$.\textsc{Append}(\texttt{NONE})\label{alqline:first_operation_ed}
\State \Call{DepthFirstTraversal}{$u_{\texttt{Ans}}$}
\State $\mathcal{A}_o$.\textsc{Append}($[\texttt{End}, 0, \texttt{NONE}]$)
\State \Return $\mathcal{A}_o$, $\mathcal{A}_v$, $\mathcal{A}_e$
\end{algorithmic}
\end{algorithm}

The total process is regarded as a traversal on the gold query graph $\mathcal{G}_q^+$. The traversal is started from vertex $v_{\texttt{Ans}}$, which is the vertex of class \texttt{Ans} because each query graph must have a vertex denoting the answer. 

At the beginning of traversal, $[\texttt{Ans}, 0, \texttt{NONE}]$, the arguments of the first operation, \textit{AddVertex}, are pushed into $\mathcal{A}_o$ (line \ref{alqline:first_operation_st} in Algorithm \ref{alg:signal_building}) and \texttt{NONE}, the instance of $v_{\texttt{Ans}}$, is pushed into $\mathcal{A}_v$ (line \ref{alqline:first_operation_ed} in Algorithm \ref{alg:signal_building}). 

Thereafter, each vertex of $\mathcal{G}_q^+$ will be visited by the depth-first traversal. Suppose that $u$ is the vertex currently being visited and its neighbor vertex $v$ along an outgoing edge $e^+=\left \langle u, v \right \rangle$ is the next vertex to be visited. If regarding the step from $u$ to $v$ as an expansion process that adds $v$ to connect to $u$, the step includes three \textit{Outlining} operations, namely, \textit{AddVertex}, \textit{SelectVertex}, and \textit{AddEdge}. Three groups of arguments are $[c_v, s_{\delta}, \hat{v}]$, $[u]$, and $[c_{e^+}, \hat{e}^+]$, respectively (line \ref{algline:outlining_st}-\ref{algline:outlining_ed}). Here, $c_v$ and $c_{e^+}$ are the class labels, $s_v$ denotes the segment number of $v$, and $\hat{v}$ and $\hat{e}^+$ denote the vertex and edge to be copied by $v$ and $e^+$, respectively. If there is no vertex (edge) to copy, $\hat{v} = \texttt{NONE}$ ($\hat{e}^+ = \texttt{NONE}$). In addition, the two \textit{Filling} operations, \textit{FillVertex} and \textit{FillEdge}  can be obtained according to the \textit{AddVertex} and \textit{AddEdge}. The two arguments are $i_v$, and $i_{e^+}$, respectively (line \ref{algline:filling_st}-\ref{algline:filling_ed}). For the traversal step along the incoming edge $e^-=\left \langle v, u \right \rangle$, the process is similar (line \ref{algline:incoming_st}-\ref{algline:incoming_ed}). After the arguments of this step are obtained, the traversal continues from $v$.

Once all vertices are visited, the traversal is completed. Finally, a group of arguments $[\texttt{End}, 0, \texttt{NONE}]$ is pushed into $\mathcal{A}_o$ and it denotes the signal of ending. \textit{Outlining}.


\end{document}